\documentclass{article}

\usepackage[nonatbib,preprint]{neurips_2021}

\usepackage[utf8]{inputenc} 
\usepackage[T1]{fontenc}    
\usepackage{hyperref}       
\usepackage{url}            
\usepackage{booktabs}       
\usepackage{amsfonts}       
\usepackage{nicefrac}       
\usepackage{microtype}      

\usepackage{times}
\usepackage{epsfig}
\usepackage{graphicx}
\usepackage{amsmath}
\usepackage{amssymb}
\usepackage{wrapfig}

\usepackage[ruled,vlined]{algorithm2e}

\usepackage{amsthm}
\usepackage{bbm}
\newtheorem{definition}{Definition}

\newtheorem{proposition}{Proposition}
\newtheorem{assumption}{Assumption}
\newtheorem{claim}{Claim}
\newtheorem{lemma}{Lemma}
\usepackage{xcolor}
\usepackage{subcaption}
\usepackage{enumitem}
\newcommand{\specialcell}[2][c]{%
  \begin{tabular}[#1]{@{}c@{}}#2\end{tabular}}

\newcommand*\tcircle[1]{%
  \raisebox{-0.5pt}{%
    \textcircled{\fontsize{7pt}{0}\fontfamily{phv}\selectfont #1}%
  }%
}

\usepackage{wrapfig}

\usepackage{multirow}

\begin{document}

\title{Universum GANs: Improving GANs through contradictions}

\author{Sauptik Dhar\thanks{equal contribution} \and Javad Heydari$^*$ \and Samarth Tripathi \and Unmesh Kurup \and Mohak Shah\\
America Research Lab, LG Electronics \\
5150 Great America Pkwy, Santa Clara, CA, USA\\
{\tt\small \{sauptik.dhar, javad.heydari, samarth.tripathi, unmesh.kurup, mohak.shah\}@lge.com}
}

\maketitle

\begin{abstract}
 
Limited availability of labeled-data makes any supervised learning problem challenging. Alternative learning settings like semi-supervised and universum learning alleviate the dependency on labeled data, but still require a large amount of unlabeled data, which may be unavailable or expensive to acquire. GAN-based  data generation methods have recently shown promise by generating synthetic samples to improve learning. However, most existing GAN based approaches either provide poor discriminator performance under limited labeled data settings; or results in low quality generated data. In this paper, we propose a Universum GAN game which provides improved discriminator accuracy under limited data settings, while generating high quality realistic data. We further propose an evolving discriminator loss which improves its convergence and generalization performance. We derive the theoretical guarantees and provide empirical results in support of our approach.

\end{abstract}

\section{Introduction} \label{sect:intro}
Training deep learning algorithms under inductive settings is highly data intensive. This severely limits the adoption of these algorithms for domains such as healthcare, autonomous driving, and prognostics and  health management, that are challenged in terms of labeled data availability. In such domains, labeling very large quantities of data is either extremely expensive, or entirely prohibitive due to the manual effort required. To alleviate this, researchers have adopted alternative learning paradigms including semi-supervised~\cite{ouali2020overview}, universum~\cite{zhang2017universum,dhar2021doc3}, transductive~\cite{elezi2018transductive,shi2018transductive}  learning,  etc., to train deep learning models. Such paradigms aim to harness the information available in additional unlabeled data sources. When unlabeled data are not naturally available, synthetic samples are generated using \textit{a priori} domain information~\cite{vargas2019robustness,vapnik06,chapelle2009semi}. A more recent line of work generates additional synthetic data using GANs to boost the discriminator's performance trained under a semi-supervised learning paradigm. For instance, \cite{salimans2016improved} adopts a feature matching loss for the generator and utilizes the generated synthetic data with some (additionally) available unlabeled data to improve the discriminator performance through semi-supervised learning. \cite{dai2017good} modifies the GAN game and adopts a complimentary generator which better detects the low-density boundaries of the labeled data distribution. The semi-supervised learning based trained discriminator using these generated data and some additional unlabeled data is shown to provide better accuracies. Here, although the trained discriminator exhibits improved generalization; the generated data does not mimic the training data distribution. A more computationally intensive approach adopts the Triple GAN architecture~\cite{li2017triple}, which includes another classifier player in the two-player GAN formulation. In that setting, the generator and classifier learns the conditional distributions between input and labels, while the discriminator learns to classify fake input-label pairs. Improving upon the notion of having an additional classifier (player), ~\cite{kavalerov2021multi} rather proposes to maintain a two-player game with an auxiliary classifier term added to both discriminator and generator losses. Here the authors target to improve upon conditional wasserstein GAN (W-GAN) games by adding auxiliary multiclass Crammer and Singer hinge losses to both discriminator and generator. Finally, \cite{miyato2018virtual} adopts an alternative learning paradigm through virtual adversarial training (VAT), which smooths the output distribution of the classifier by generating carefully designed adversarial samples while assigning virtual labels to unlabeled data. For all these approaches the major gain comes from an additionally available unlabeled data.

In this paper, we consider the scenario where \textit{no additional unlabeled samples are available}. We demonstrate how our proposed approach leverages only the GAN-generated data to improve generalization while safeguarding against mode collapse compared to Feature Matching FM-GAN~\cite{salimans2016improved}, and generating more realistic synthetic data compared to Complimentary-GAN (C-GAN) \cite{dai2017good}. Our main idea pivots around training the discriminator under the universum learning setting~\cite{vapnik06,dhar2019}. Also evolving the discriminator loss from a universum to semi-supervised setting can yield further gains in discriminator generalization. The main contributions of this paper are,  
\begin{enumerate}[nosep,leftmargin=*]
\item We propose to train the discriminator under universum setting (in Section \ref{sec_UGAN}) rather than semi-supervised settings \cite{salimans2016improved,dai2017good}, and propose a generic universum GAN (U-GAN) game in eq. \eqref{eq_gan_play1}, \eqref{eq_gan_play2}. We provide the theoretical analysis of U-GAN's consistency and exemplify it for multiclass Hinge loss in \eqref{eq_ugan_hinge}. 
\item Next, we motivate evolving the discriminator loss from universum to semi-supervised setting to propose the new Evolving GAN algorithm in Section~\ref{sec_Egan}. The proposed evolving mechanism further improves upon U-GAN's discriminator generalization. We also derive a unified loss which can evolve the discriminator loss seamlessly from universum to semi-supervised in Section~\ref{sec:unified_loss}.
\item Finally, we empirically demonstrate the effectiveness of our proposed approach in Section~\ref{sec_results}.
\end{enumerate}

The paper is organized as follows. Section \ref{sec:preliminaries} provides preliminaries on the different learning settings and exemplifies  C\&S  hinge loss \cite{crammer02} under these settings. A unified loss to solve both universum and semi-supervised C\&S hinge is also provided. Section \ref{sec_UGAN} introduces the new Universum GAN game, and provides the theoretical analysis on its consistency. Section \ref{sec_Egan} motivates evolving the learning paradigm of the discriminator loss from universum $\rightarrow$ semi-supervised setting and proposes the new evolving GAN algorithm in Algorithm~\ref{algo:evolving_gan}. Section \ref{sec_results} provides the empirical results. Section \ref{sec_conc} provides the conclusions. \begin{wrapfigure}[12]{r}{0.35\textwidth}
  \begin{center}
    \includegraphics[width=0.35\textwidth]{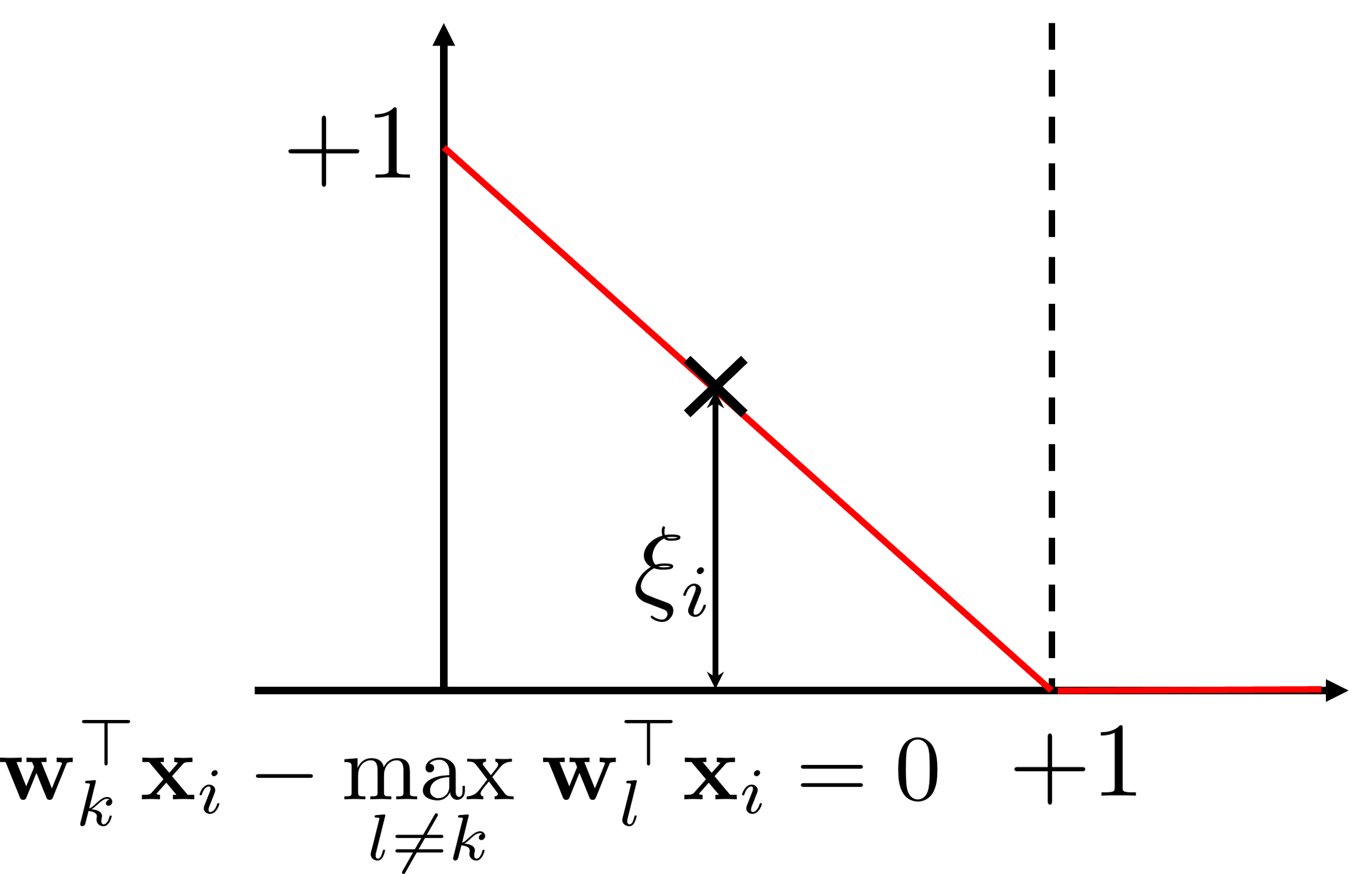}
  \end{center} 
  \caption{C \& S Hinge loss under inductive settings. Sample $(\mathbf{x}_i,y_i)$ lying inside the margin is linearly penalized using slack variable $\xi_i$.} \label{fig_hinge} 
\end{wrapfigure} 

\section{Preliminaries on Learning Settings} 
\label{sec:preliminaries} 

We first introduce the learning settings that  will be used in this paper  and exemplify them with the  C\&S-hinge loss \cite{crammer02}. 

\subsection{Inductive Learning}

As the most widely used learning setting in machine learning and deep learning, it aims to estimate a model using the labeled training data to predict on future test samples. The mathematical formalization of this setting is provided below.
\begin{definition} \textbf{(Inductive Learning)} \label{def_inductive}
Given i.i.d training samples $\mathcal{T}=(\mathbf{x}_i,y_i)_{i=1}^n \sim \mathcal{D}_{\mathcal{X}}^n \times \mathcal{D}_{\mathcal{Y}}^n$, with $\mathbf{x} \in \mathcal{X} \subseteq \Re^d$ and $y \in \mathcal{Y} = \{1,\ldots, L\}$, estimate a hypothesis $h^*:\mathcal{X} \rightarrow \mathcal{Y}$ from a hypothesis class $\mathcal{H}$ which minimizes, 
\begin{flalign} \label{eq_inductive}
 \underset{h \in \mathcal{H}}{\text{inf}} \; \mathbb{E}_{\mathcal{D}_{\mathcal{T}}}[\mathbbm{1}_{(y \neq h(\mathbf{x}))}]
\end{flalign}
where, $\mathbb{E}_{\mathcal{D}_{\mathcal{T}}}$ is the  expectation under training distribution $\mathcal{D}_{\mathcal{T}}$, and $\mathbbm{1}_{(\cdot)}$ is the indicator function.
\end{definition} 
\noindent A popular approach is to estimate a multi-valued function $\mathbf{f} = [f_1, \ldots, f_L]$ and use the decision rule, 
\begin{align} 
\label{eq_dec_rule}
 h(\mathbf{x}) \; \left\{
\begin{array}{l l}
   =  k & \text{if } f_k(\mathbf{x}) > f_\ell(\mathbf{x}) \;;  \forall \ell \neq k  \\
   \neq [1,\ldots, L] &  \text{else}
\end{array}\right. \ 
\end{align}

\noindent There are several existing algorithms to estimate this multi valued function. The C\&S hinge is one widely used approach which adopts a margin based loss function shown below,
\begin{flalign} 
\label{eq_cs_loss}
&\min_{\mathbf{w}_1 \ldots \mathbf{w}_L ,\boldsymbol\xi} \sum\limits_{i=1}^n \xi_{i} \ , \ \text{s.t.} \; \xi_i = \max_{k \in \mathcal{Y}} \; \{ 1-\delta_{ik} + (\mathbf{w}_k^T-\mathbf{w}_{y_i}^T)\mathbf{x}_i \}
\end{flalign}
where, $\delta_{i\ell}=\mathbbm{1}_{(y_i=\ell)}$. Throughout we use linear parameterization $f_k(\mathbf{x}) = \mathbf{w}_k^{\top} \mathbf{x}$ for simplicity. Here, any training sample $(\mathbf{x}_i,y_i)$ lying inside the margin $+1$ is linearly penalized using a slack variable $\xi_i$ (see Fig \ref{fig_hinge}). The C\&S-hinge loss minimizes the approximation error while keeping the estimation error small compared to other multi-class loss alternatives \cite{daniely2012multiclass}, and presents itself as a reliable choice for limited data settings. However, for high dimensional limited labeled data problems, even such advanced hinge-based loss function may fail to provide desired generalization. This motivates the need for novel learning settings discussed next.

\subsection{Semi-Supervised Learning}
Semi-supervised learning is a widely used advanced learning setting. Here, in addition to labeled training data we are also given with unlabeled samples which follow a similar distribution as the labeled data. The goal here is to leverage the additional unlabeled data to improve the test time accuracy. The setting is formalized as,
\begin{definition} \textbf{(Semi-Supervised Learning)}  \label{def_semisetting}
Given $n$ i.i.d training samples $\mathcal{T}$,
and additional $m$ unlabeled samples $\mathcal{U} = (\mathbf{x}_{i^\prime}^{*})_{i^\prime=1}^m \sim \mathcal{D}_{\mathcal{X}}^m$ with $\mathbf{x}^{*} \in \mathcal{X}_{U}^* \subseteq \Re^d$, estimate $h^*:\mathcal{X} \rightarrow \mathcal{Y}$ from $\mathcal{H}$ which solves~\eqref{eq_inductive} 
\end{definition}

A popular C\&S hinge extension under this setting follows \cite{zien2007transductive}:
\begin{flalign} \label{eq_cs_transductive_loss}
& \min_{\mathbf{w}_1 \ldots \mathbf{w}_L ,\boldsymbol\xi} \quad \quad  \sum\limits_{i=1}^n \xi_{i} +  C_U \sum\limits_{i^\prime = 1}^{m} \zeta_{i^\prime} \\
&\text{s.t.} \quad \xi_i = \max_{k \in \mathcal{Y}} \; \{ 1-\delta_{ik} + \mathbf{w}_k^T\mathbf{x}_i - \mathbf{w}_{y_i}^T\mathbf{x}_i \}  \quad \exists y_{i^\prime}^* : \zeta_{i^\prime} = \max_{k \in \mathcal{Y}} \; \{ 1-\delta_{i^{\prime}k} + \mathbf{w}_k^T\mathbf{x}_{i^{\prime}}^* - \mathbf{w}_{y_{i^\prime}^* }^T\mathbf{x}_{i^\prime}^* \} && \nonumber
\end{flalign}
\noindent Here, in addition to the traditional C\&S hinge loss on the labeled data, we use a similar margin based loss on the unlabeled data. However, different from the labeled counterpart; we expect to minimize the C\&S hinge loss for \textit{some} labeling $y_{i^\prime}^*$ on the unlabeled data. \begin{wrapfigure}[17]{r}{0.45\textwidth}
  \begin{center}
    \includegraphics[width=0.45\textwidth]{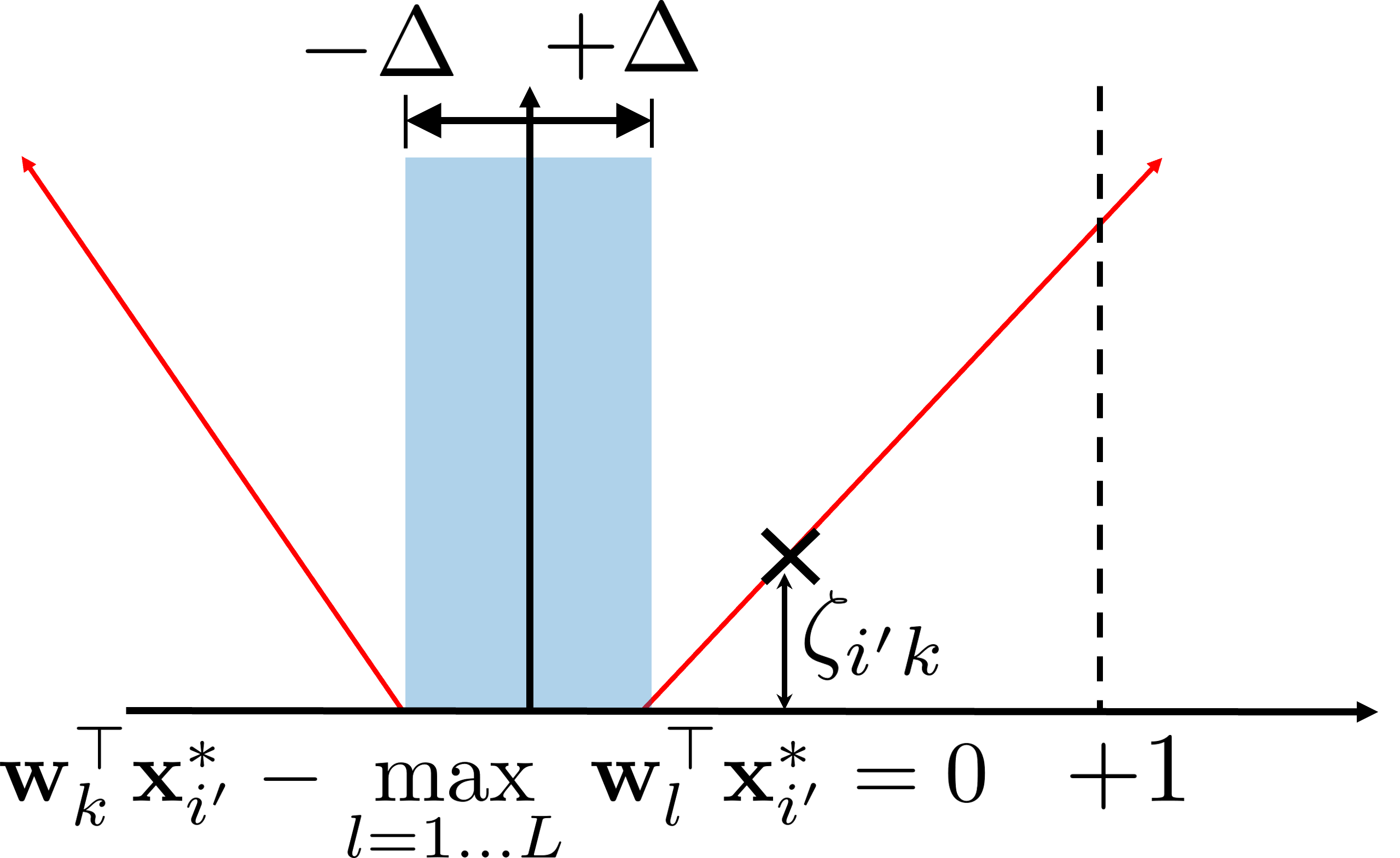}
  \end{center}
  \caption{Universum loss for $k^{th}$ class decision boundary. Universum samples $(\mathbf{x}_{i^{\prime}}^{*})$ lying outside the $\pm \Delta$ -insensitive zone is linearly penalized using slack variable $\zeta_{i^{\prime}k}$.} \label{univ_hinge}
\end{wrapfigure}


\subsection{Universum a.k.a Contradiction Learning}
Another advanced learning setting is the universum a.k.a contradiction learning setting. Here, in addition to the labeled training data we are also given with unlabeled universum samples which are known not to belong to any of the classes in the training data. For example, if the goal of learning is to discriminate between handwritten digits (0, 1, 2,...,9), one can
introduce additional ‘knowledge’ in the form of handwritten letters (A, B, C, ... ,Z). These examples from the universum contain certain information (e.g., handwriting styles) but they cannot be assigned to any of the classes (0 to 9). Further, the universum samples do not have the same distribution as labeled training samples. Learning under this setting can be formalized as below,
\begin{definition} \textbf{(Universum Learning)}  \label{def_univsetting}
Given $n$ i.i.d training samples $\mathcal{T}$,
and additional $m$ unlabeled universum samples $\mathcal{U} = (\mathbf{x}_{i^\prime}^{*})_{i^\prime=1}^m \sim \mathcal{D}_{\mathcal{U}}$ with $\mathbf{x}^{*} \in \mathcal{X}_{U}^* \subseteq \Re^d$, estimate $h^*:\mathcal{X} \rightarrow \mathcal{Y}$ from hypothesis class $\mathcal{H}$ which, in addition to solving~\eqref{eq_inductive}, obtains maximum contradiction on universum samples i.e., it is the solution to 
\begin{flalign} \label{eq_contradition}
\underset{h \in \mathcal{H}}{\text{sup}} \; \mathbb{P}_{\mathcal{D}_{\mathcal{U}}}[\mathbf{x}^* & \notin \text{any class}] = \underset{h \in \mathcal{H}}{\text{sup}} \; \mathbb{E}_{\mathcal{D}_{\mathcal{U}}}[\mathbbm{1}_{ \lbrace \bigcap\limits_{k \in \{ 1,\ldots ,L \}} h(\mathbf{x}^*) \neq k \rbrace}] \ .
\end{flalign}
where, $\mathcal{D}_{\mathcal{U}}$ is the universum distribution, $\mathbb{P}_{\mathcal{D}_{\mathcal{U}}}(\cdot)$ and $\mathbb{E}_{\mathcal{D}_{\mathcal{U}}}(\cdot)$ are the probability measure and expectation under the universum distribution, respectively, and $\mathcal{X}_{U}^{*}$ is the domain of universum data.
\end{definition}
Recently \cite{dhar2019} proposed a C\&S Hinge extension under universum setting. Their approach relies on the following proposition, 
\begin{proposition}  \label{prop_max_contradiction}
For the C\&S formulation in \eqref{eq_cs_loss} and the corresponding decision rule in \eqref{eq_dec_rule}, maximum contradiction on universum samples $\mathbf{x}^* \in \mathcal{U}$ can be achieved when,
\begin{flalign} \label{max_contradiction}
|\mathbf{w}_k^T \mathbf{x}^{*}-\max_{\ell\in \mathcal{Y}}\ \mathbf{w}_\ell^T \mathbf{x}^{*}| = 0\ ; \; \forall k \in \mathcal{Y}
\end{flalign}
\end{proposition}
\noindent In practice the constraint in \eqref{max_contradiction} is relaxed using a $\pm \Delta$ - insensitive loss to solve,
\begin{flalign} \label{eq_cs_universum}
    & \underset{\mathbf{w}_1 \ldots \mathbf{w}_L ,\boldsymbol\xi,\boldsymbol\zeta}{\text{min}}  \quad 
    \sum\limits_{i=1}^n \xi_{i} + C_U\sum\limits_{i^\prime =1}^m \sum\limits_{k=1}^L \zeta_{i^\prime k}  \quad  \forall i\in \{1,\dots, n\} \quad \forall i^\prime \in \{1,\dots, m\} && \\
    & \text{s.t.} \quad \xi_i = \underset{k \in \mathcal{Y}}{\text{max}} \; \{ 1-\delta_{ik} + \mathbf{w}_k^T\mathbf{x}_i - \mathbf{w}_{y_i}^T\mathbf{x}_i \} \; \text{and} \;   \zeta_{i^\prime k} = \text{max} \{|\mathbf{w}_k^T \mathbf{x}_{i^\prime}^*-\max_{\ell\in \mathcal{Y}}\mathbf{w}_l^T \mathbf{x}_{i^\prime}^*| - \Delta, 0 \} \nonumber &&
\end{flalign}
\noindent Here, for the $k^{th}$ class decision boundary the universum samples that lie outside the
$\Delta-$insensitive zone are linearly penalized using the slack variables $\zeta_{i^{\prime}k}$ (see Fig \ref{univ_hinge}). The user-defined parameters $C_U \geq 0$ control the trade-off between the margin-error on training samples, and the contradictions (samples lying outside $\pm \Delta$ zone) on the universum samples.

\subsection{Unified Loss for Solving C\&S Hinge Loss Under Different Learning Settings}
\label{sec:unified_loss}

In this paper, we introduce a unified loss to solve both the optimization problems in eqs. \eqref{eq_cs_transductive_loss} and \eqref{eq_cs_universum}. This follows from a similar transformation in  Proposition $3$ of~\cite{dhar2019},  
\begin{definition} \label{def_transformation} \textbf{(Transformation)} For each unlabeled sample $\mathbf{x}_{i^\prime}^*$ we create $L$ artificial samples belonging to all classes i.e. $(\mathbf{x}_{i^\prime}^{*},y_{i^\prime 1}^{*}=1),\ldots, (\mathbf{x}_{i^\prime}^{*},y_{i^\prime L}^{*}=L)$.
\end{definition}
\noindent With the above transformation we solve, 
\begin{flalign} \label{eq_cs_unified_loss}
    &\underset{\mathbf{w}_1 \ldots \mathbf{w}_L ,\boldsymbol\xi}{\text{min}} \quad
    \sum\limits_{i=1}^n \xi_{i} +  C_U \sum\limits_{i=n+1}^{n+mL} \xi_{i} && \\
    &\text{s.t.} \;
    \xi_i = \underset{k \in \mathcal{Y}}{\text{max}} \; \{ 1-\delta_{ik} + \mathbf{w}_k^T\mathbf{x}_i - \mathbf{w}_{y_i}^T\mathbf{x}_i \} , \quad  i = 1 \ldots n && \nonumber\\ 
    &\xi_i = \psi_{\epsilon}\Big(\underset{k \in \mathcal{Y}}{\text{max}} \; \{ \epsilon (1-\delta_{ik}) + \mathbf{w}_k^T\mathbf{x}_{i} - \mathbf{w}_{y_{i}}^T\mathbf{x}_i \}\Big),\;  i = n+1 \ldots n+mL && \nonumber
\end{flalign}
Appropriately selecting the $\psi_{\epsilon}(\cdot)$ and $\epsilon$ provides us the desired solutions for both \eqref{eq_cs_transductive_loss} and \eqref{eq_cs_universum}.
\begin{proposition} \label{prop_tran_usvm}
Solving \eqref{eq_cs_unified_loss} with $\epsilon = -\Delta$ and $\psi_{\epsilon}(x) = x$ provides the solution to \eqref{eq_cs_universum}.
\end{proposition} 

\begin{proposition} \label{prop_tran_tsvm}
Solving \eqref{eq_cs_unified_loss} with $\epsilon = 1$ and $\psi_{\epsilon}(x) = \text{min}\{x,\epsilon\}$ provides a solution to \eqref{eq_cs_transductive_loss}.
\end{proposition} 
\noindent The advantages of this singular framework are two-fold.
\begin{itemize}[nosep,left=0pt]
    \item[--] First, we can solve either of the formulations \eqref{eq_cs_transductive_loss} or \eqref{eq_cs_universum} using \eqref{eq_cs_unified_loss} by carefully tuning $\epsilon$ and $\psi(\cdot)$. In fact, this also provides us with the framework to transition the learning setting from universum to semi-supervised (i.e. with $\epsilon = -\Delta \rightarrow 1 $) when the data distribution of the unlabeled samples change from being contradictions (i.e. $\mathbf{x}^* \in \mathcal{X}^*_U$) to being compliant (i.e. $\mathbf{x}^* \in \mathcal{X}$). This will be a very useful tool for the evolving GAN game later introduced in section \ref{sec_Egan}.
    \item[--] Second, the Propositions \eqref{prop_tran_usvm} and \eqref{eq_cs_universum} can harness the advanced optimization techniques used to solve the standard C\&S hinge loss. This property has already been established for universum settings \cite{dhar2019}. For the semi-supervised setting, prior solvers \cite{zien2007transductive,balamurugan2013large,selvaraj2012extension} to \eqref{eq_cs_transductive_loss} use a switching algorithm which incurs significant computation complexity. Through Def. \eqref{def_transformation} we can avoid such switching algorithms and still attain similar performance results (see results in Appendix~\ref{app_TSVM_baselines}).   
\end{itemize}

\section{Universum GAN (U-GAN)} \label{sec_UGAN} 
\begin{wrapfigure}{r}{0.45\textwidth}
  \begin{center}
    \includegraphics[width=0.45\textwidth]{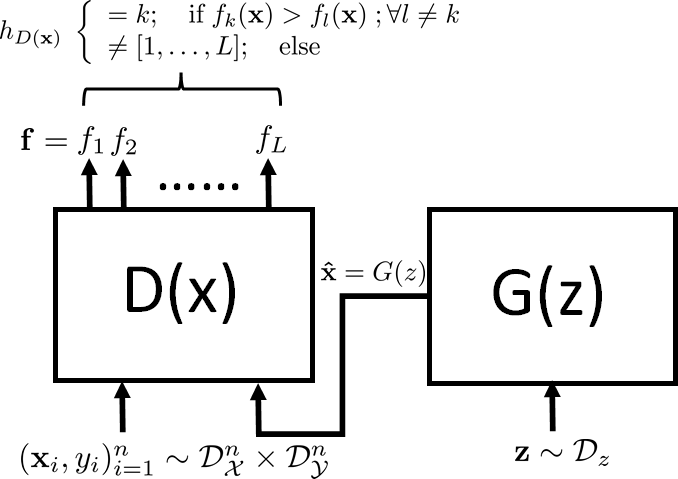}
  \end{center}
  \caption{Two player GAN game.}
  \label{fig:sgan_game}
\end{wrapfigure}

With the preliminaries on different learning settings in place, and a unified loss to solve the C\&S loss for all these settings; next we introduce the new universum GAN game  (see Fig.~\ref{fig:sgan_game}), 
\begin{flalign}
\text{Player 1:} \quad & \underset{D}{\text{max}} \; L_D = \underset{XY}{\mathbbm{E}}[\mathbbm{1}_{(y=h_{D(\mathbf{x})})}] +  C_G\; \underset{z}{\mathbbm{E}}[\mathbbm{1}_{h_{D(G(z))} \notin \mathcal{Y}}] \label{eq_gan_play1} && \\
\text{Player 2:} \quad & \underset{G}{\text{max}} \; L_G = \underset{z}{\mathbbm{E}}[\mathbbm{1}_{h_{D(G(z))} \in \mathcal{Y}}]  \label{eq_gan_play2} &&
\end{flalign} 
$D$ = Discriminator, $G$ = Generator, $h_D$ = Decision rule as in \eqref{eq_dec_rule} induced by $D$. Note that the GAN game in   \eqref{eq_gan_play1} and \eqref{eq_gan_play2} has the same intuition as the original semi-supervised GAN \cite{salimans2016improved}. That is, Player 1 estimates a discriminator that explains the training samples (classes $1$ through $L$) while simultaneously identifying the generated samples to not belong to any class. On the other hand, Player 2 confuses the discriminator by generating samples as belonging to one of the discriminator classes. However, different from~\cite{salimans2016improved} we do not assign all the generated samples to belong to one separate class (say $L+1$). Rather, we utilize the universum setting and treat the generated samples as contradictions. This is a more desirable setting, as it does not make an overgeneralized assumption that \textit{all the generated samples belong to the same class $L+1$}. Next we provide the theoretical justification behind our formulation in Proposition \ref{prop_UGAN}.  Here we use a discriminator that estimates a multi-valued function $f = [f_1,\ldots,f_L]$ and the decision rule $h$ as in~\eqref{eq_dec_rule}. To simplify the proof we use the following assumption,
\begin{assumption} \label{assumption_bayes}(Realizability) There exist a measurable function $h^*$ that achieves zero Bayes Risk on the training data distribution $R(h^*) = \mathbbm{E}_{(\mathbf{x},y) \sim \mathcal{D}_{\mathcal{X}} \times \mathcal{D}_{\mathcal{Y}}}[\mathbbm{1}_{y=h^*(\mathbf{x})}] = 0$
\end{assumption}
\begin{proposition} \label{prop_UGAN} (Consistency)
Under assumption \ref{assumption_bayes} $ \exists \; C_G \leq 1$ such that the optimal $(D^*,G^*)$ that solves the GAN game in~\eqref{eq_gan_play1} and~\eqref{eq_gan_play2} satisfies the following,
\begin{enumerate}[nosep, label=(\roman*)]
    \item $D^*$ achieves Bayes Risk on $(\mathbf{x},y) \sim \mathcal{D}_{\mathcal{X}} \times \mathcal{D}_{\mathcal{Y}}$, i.e., $R(h_{D^*}) = 0$.
    \item The support of the generated data (i.e. support of \; $\mathbbm{P}_{G^*}$) is contained in $\mathcal{X}$.
\end{enumerate}
\end{proposition}
The above proposition establishes that the 2-player game in~\eqref{eq_gan_play1} and~\eqref{eq_gan_play2} indeed generates samples from the training data distribution; while achieving the best possible generalization performance for the discriminator. However, the proposition holds under a strong assumption~\ref{assumption_bayes}. This assumption provides us with a mathematical construct that simplifies the proof significantly. However, we argue that the proposition \ref{prop_UGAN} holds even without the realizability assumption.
\begin{claim} \label{claim_UGAN}
For appropriately selected $C_G$, the proposition \eqref{prop_UGAN} holds without assumption~\ref{assumption_bayes}.
\end{claim}

The Proposition \ref{prop_UGAN} provides the theoretical consistency for the U-GAN formulation, generally missing for most existing semi-supervised GAN formulations~\cite{salimans2016improved,kavalerov2021multi}. Note however, the loss functions in~\eqref{eq_gan_play1} and~\eqref{eq_gan_play2} are not differentiable. In this work, we use the C\&S hinge loss as a dominating surrogate for the discriminator and generator loss. That is, we use the universum loss \eqref{eq_cs_universum} for the discriminator, and the  unlabeled component of semi-supervised loss \eqref{eq_cs_transductive_loss} for generator. We also add the feature matching loss to the generator. The final U-GAN game is given as,
\begin{flalign} \label{eq_ugan_hinge}
&\mathbf{L_D} = \sum\limits_{i=1}^n \xi_{i} + C_U\sum\limits_{i^\prime =1}^m \sum\limits_{k=1}^L \zeta_{i^\prime k}  \; \text{and} \;  \mathbf{L_G} = \hat{C}_U\sum\limits_{i^\prime = 1}^{m} \hat{\zeta}_{i^\prime} + ||\mathbb{E}[ \boldsymbol\phi(G(\mathbf{n};\boldsymbol{\theta}))]-\mathbb{E}[\boldsymbol\phi(\mathbf{x})]||_1  && \\
&\text{s.t.} \;  \xi_i = \underset{k \in \mathcal{Y}}{\text{max}} \; \{ 1-\delta_{ik} + \mathbf{w}_k^T\mathbf{z}_i - \mathbf{w}_{y_i}^T\mathbf{z}_i \}; \quad  \forall  i=1\ldots n; \; \forall i^\prime = 1 \ldots m.  && \nonumber \\
& \zeta_{i^\prime k} = \text{max} \{|\mathbf{w}_k^T \mathbf{z}_{i^\prime}^*-\max_{\ell\in \mathcal{Y}}\mathbf{w}_l^T \mathbf{z}_{i^\prime}^*| - \Delta, 0 \}; \quad  \exists y_{i^\prime}^* : \hat{\zeta}_{i^\prime} = \max_{k \in \mathcal{Y}} \; \{ 1-\delta_{i^{\prime}k} + \mathbf{w}_k^T\mathbf{z}_{i^{\prime}}^* - \mathbf{w}_{y_{i^\prime}^* }^T\mathbf{z}_{i^\prime}^* \}    \nonumber &&
\end{flalign}
\noindent Here, $f_k(\mathbf{x}) = \mathbf{w}_k^{\top}\mathbf{z}$, $\mathbf{z} = \boldsymbol\phi(\mathbf{x}) \quad \mathbf{z}_{i^\prime}^* = \boldsymbol\phi(\mathbf{x}_{i^\prime}^*)$, where $\phi$ is the feature map induced by the discriminator network; and generator $G(\mathbf{n};\boldsymbol{\theta})$ is parameterized with $\boldsymbol{\theta}$, with input  noise $\mathbf{n}$. Note that, formulation \eqref{eq_ugan_hinge} address a previous shortcoming identified for multiclass conditional GANs that \textit{classification and discrimination should be left as auxilliary tasks} \cite{kavalerov2021multi}. Prop. \ref{prop_max_contradiction} shows how $\mathbf{L_D}$ in \eqref{eq_ugan_hinge} simultaneously targets good classification on labeled data, while discriminating between real vs. fake (contradiction / universum) samples. Further, Prop. \ref{prop_UGAN} guarantees the consistency of the GAN game using such a loss function.    

From a practical perspective, an immediate advantage of the U-GAN in \eqref{eq_ugan_hinge} compared to semi-supervised GAN~\cite{salimans2016improved} is that, it can provide an implicit regularization to increase the entropy of the predicted labels on generated samples. This intuition follows from the empirical results reported in~\cite{dhar2019}. Through the histogram of projections (HOP) visualization \cite{dhar2019} demonstrated how universum model results to higher entropy on their predicted labels compared to inductive settings. In fact, for binary problems, \cite{sinz2007priori} derives the connection between hard-margin universum and the maximum entropy solution. This maximum entropy property is highly desirable for GAN games as it alleviates the mode-collapse problem. Empirical results on this implicit regularization is provided in Section~\ref{sec_results}. 

\section{Evolving GAN (E-GAN): From Contradictions to Compliance} \label{sec_Egan}


\begin{wrapfigure}{r}{.5\textwidth}
\begin{minipage}{\linewidth}
\centering
\subcaptionbox{Model performance.}
{\includegraphics[width=5.5cm]{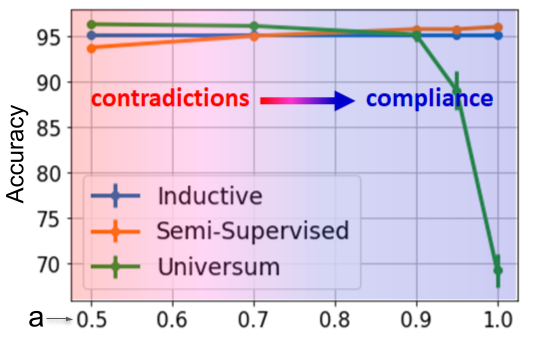}}\\
\subcaptionbox{a=.5}
{\includegraphics[width=1.1cm]{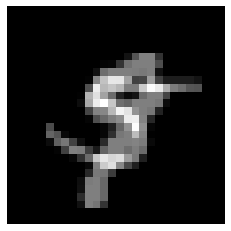}}
\subcaptionbox{a=.7}
{\includegraphics[width=1.1cm]{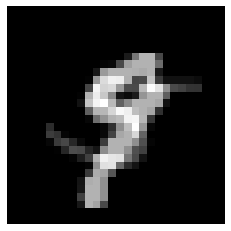}}
\subcaptionbox{a=.9}
{\includegraphics[width=1.1cm]{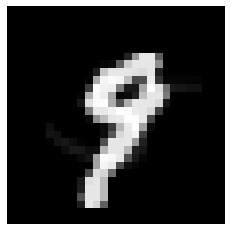}} 
\subcaptionbox{a=.95}
{\includegraphics[width=1.1cm]{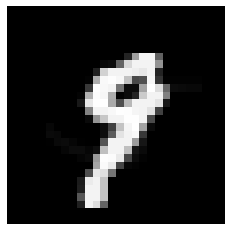}} 
\subcaptionbox{a=1.0}
{\includegraphics[width=1.1cm]{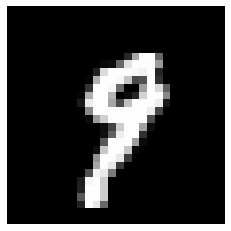}}
\end{minipage}
\caption{(a) Model performance under different learning settings with changing distribution of the unlabeled data generated by mixing randomly selected training images with ratios $a = 0.5 \rightarrow 1.0$ (b)-(f) Example images generated by mixing randomly selected digits `5' and `9' with ratios $a = 0.5 \rightarrow 1.0$.} \label{fig:mixdistribution}
\end{wrapfigure}


Although U-GAN admits a consistent solution (see Prop. \ref{prop_UGAN})  and guarantees advantages on the mode  collapse problem seen for semi-supervised GAN \cite{salimans2016improved}; it still has the same caveats as discussed in \cite{dai2017good}.  Rightly so, since close to convergence the generated data follows a very similar distribution as the training data. This violates the Universum assumption in Definintion \ref{def_univsetting}; where the generated (universum) samples should act as contradictions and results in sub-optimal discriminator performance. Rather, a more apt setting for the discriminator when the generated data is in {\em compliance} with the training samples, is the semi-supervised setting in Def. \eqref{def_semisetting}. This can be better explained using the synthetic example in Fig.~\ref{fig:mixdistribution}, which shows the performance of a linear model trained under different learning settings, i.e., inductive eq. ~\eqref{eq_cs_loss}, semi-supervised eq. ~\eqref{eq_cs_transductive_loss} and universum eq.~\eqref{eq_cs_universum} using the standard MNIST data~\cite{lecun-mnisthandwrittendigit-2010}. Here, the goal is to build a multiclass `0'--`9' digit classifier using 50000 training samples to predict on 10000 test samples. Here, to simulate the changing distributions of the unlabeled data we randomly select any two training images $(\mathbf{x}_i,y_i), (\mathbf{x}_j,y_j)$ and perform a weighted average $\mathbf{x}^* = a \mathbf{x}_i + (1-a) \mathbf{x}_j$, with (mixing ratio $a \in [0.5,1.0]$) to generate an unlabeled sample. Example of such a generated universum sample using a randomly selected digit `5' and `9' image for different mixing ratios is shown in Fig. \ref{fig:mixdistribution} (b) - (f). As seen in Fig. \ref{fig:mixdistribution} (a) universum outperforms the other approaches when the generated data act as contradictions a = 0.5 (i.e. neither `5' or `9'). However, as the mixing ratio increases $a > 0.9$, the performance under universum learning deteriorates. Rightly so, since with $a>0.9$ the generated data closely resembles the training data. Training under semi-supervised setting is a more desirable choice. The main takeaway  from this example is that, as the distribution of the generated data changes from contradictions to compliance, it is favorable to evolve the discriminator loss from universum to semi-supervised setting. Doing so, may yield improved generalization performance. In this work, we adopt this intuition and evolve the discriminator loss for improved  generalization. Note that mechanisms similar to C-GAN \cite{dai2017good} could have been adopted, where we rather generate complimentary samples to boost the U-GAN's performance. However, as discussed in section \ref{sect:intro}, such an approach will result to non real-like generated samples and is contrary to our overall goal (later confirmed through results in Fig \ref{fig:gen_data}). 

\begin{wrapfigure}{L}{0.45\textwidth}
\begin{minipage}{0.45\textwidth}
\begin{algorithm}[H] \small
\SetAlgoLined
 \textbf{Initialize} Discriminator and Generator \;
 \textbf{Parameters:} \texttt{epsSet}, \texttt{evolvePeriod}, \texttt{numiter}, and $C_U$ \;
 \For{$i\gets0$ \KwTo \texttt{numiter}}{
 	Select $M$ samples from the dataset \;
    Generate $M$ samples using generator\;
    Update $\epsilon \gets \texttt{epsSet}\left[\frac{i}{\texttt{evolvePeriod}}\right]$\;
    \eIf{$\epsilon < 0$}{
      Update discriminator \eqref{eq_ugan_hinge} under universum setting\;
      Update generator in \eqref{eq_ugan_hinge}\;
    }{
      Update discriminator \eqref{eq_ugan_hinge} under semi-supervised setting\;
    }
    }
 \caption{Evolving GAN Algorithm}
 \label{algo:evolving_gan}
\end{algorithm}
\end{minipage}
\end{wrapfigure} 

To design our evolving mechanism we utilize the Propositions~\ref{prop_tran_usvm} and~\ref{prop_tran_tsvm}. This allows us to seamlessly transition from a universum learning setting to a semi-supervised setting by changing $\epsilon$ from $-\Delta$ to $1$. In this paper, we adopt this unified loss and update $\epsilon$ in a staircase fashion. Specifically, we define a set of $\epsilon$ values $\texttt{epsSet = [-0.05,-0.01,\ldots,1.0]}$, start the training process with $\epsilon=-0.05$ (universum learning), and after each $\texttt{evolvePeriod = 5000}$ iteration, we select the next value from set $\texttt{epsSet}$. Such a simple evolution routine may not be optimal, but has shown  significant performance gains in our results (see Section \ref{sec_results}). Note that, $\epsilon<0$ corresponds to universum learning setting, while $\epsilon=1$ leads to semi-supervised loss. Also for this work, we stop training the generator as the discriminator switches to semi-supervised setting. A more advanced evolution mechanism and a detailed study on optimal mechanisms for training the generator even during the semi-supervised learning phase is still an open research problem.

\section{Empirical Results} \label{sec_results}

For our experiments, we use the same network architecture for discriminator and generator as in~\cite{dai2017good}. Similar to~\cite{dai2017good}, we randomly sample $1,000$ and $4,000$ labeled data from SVHN and CIFAR-10 datasets, respectively. However, unlike~\cite{dai2017good}, we do not use any additional unlabeled data. \begin{wraptable}[16]{r}{0.6\textwidth}
    \centering
    
    \caption{Comparison with baseline methods on SVHN and CIFAR-$10$ datasets. $^*$ = original paper's results using additional unlabeled data. $\dagger$ = without using unlabeled data.} \label{tab:error}
    
    \resizebox{0.6\columnwidth}{!}{
    \begin{tabular}{lll}
        \hline 
        \addlinespace
         Method & SVHN & CIFAR-$10$ \\
         \hline
         ADGM$^*$~\cite{maaloe2016auxiliary} & $22.86$ & - \\
         SDGM$^*$~\cite{maaloe2016auxiliary} & $16.61 \pm 0.24$ & - \\
         FM$^\dagger$~\cite{salimans2016improved} & $19.65 \pm 1.74$ & $35.37 \pm 1.56$ \\
         C-GAN$^\dagger$~\cite{dai2017good} & $15.56 \pm 1.68$ & $35.60 \pm 0.78$ \\
         U-GAN$^\dagger$ (ours) & $15.04 \pm 0.77$ & $31.76 \pm 0.85$ \\
         U-GAN + PT/VI$^\dagger$ (ours) & $\mathbf{14.84 \pm 0.88}$ & $\mathbf{29.53 \pm 0.81}$ \\ \addlinespace
        \midrule
         VAT large$^\dagger$~\cite{miyato2018virtual} & $14.59 \pm 1.31$ & $19.17 \pm 0.1
         9$ \\
         U-GAN + VAT$^\dagger$ (ours) & $\mathbf{10.21 \pm 0.52}$ & $\mathbf{17.32 \pm 0.93}$ \\
         \hline
    \end{tabular}
    }

\end{wraptable}

\subsection{Effectiveness of U-GAN}\label{sec_results_UGAN_effective}
\textbf{Classification Accuracy}:  First, we compare the performance of U-GAN with the popular GAN based algorithms for limited data settings reported in \cite{dai2017good}. Table \ref{tab:error} provides the mean $\pm$ standard deviation of the classification error over 10 random partitioning of the training data. Note that, we mainly compare our approach with  FM \cite{salimans2016improved} and C-GAN \cite{dai2017good}, as these approaches also provide alternative mechanisms to improve discriminator's generalization under limited labeled data settings, by integrating a loss term associated with the generated samples. For fair comparison, we repeat the experiments for FM and C-GAN, and remove the loss terms corresponding to the unlabeled data during training. For feature matching loss, we use a randomly sampled labeled data to obtain the features' statistics. 

Table~\ref{tab:error} shows that U-GAN  outperforms both FM and C-GAN approaches. This is due to the fact that under universum setting the generated samples act as contradictions, which better constraints the search space of the optimal model. Such a behavior is inline with previous research on universum learning~\cite{vapnik06,sinz08,dhar2019}. Since the pull-away term (PT) and variational inference (VI) techniques used in~\cite{dai2017good} for increasing generator's entropy are orthogonal to the U-GAN model, we observe that they further improve the performance of U-GAN. This may be due to more diverse samples being generated by the U-GAN+VI/PT (later confirmed in Table~\ref{tab:entropy}). Finally, following \cite{dai2017good} we also provide performance comparisons with the VAT algorithm \cite{miyato2018virtual}. Note that, VAT adopts an adversarial learning setting and is orthogonal to our proposed Universum approach. In fact, combining U-GAN with VAT has compounding effect that leads to significant improvements for SVHN and CIFAR datasets.

\begin{figure*}[t]
    \centering
    \begin{subfigure}{.28\textwidth}
      \centering
      \includegraphics[width=.95\linewidth]{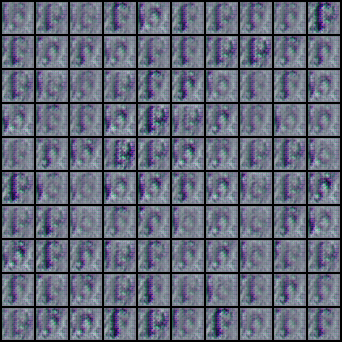}
      \caption{FM~ (SVHN)}
      \label{fig:a}
    \end{subfigure}%
    \begin{subfigure}{.28\textwidth}
      \centering
      \includegraphics[width=.95\linewidth]{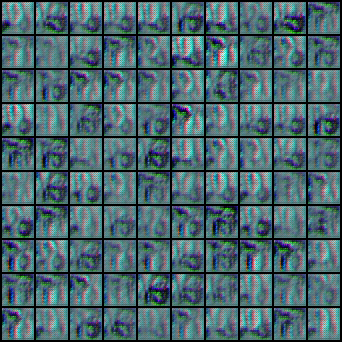}
      \caption{C-GAN~ (SVHN)}
      \label{fig:b}
    \end{subfigure}%
    \begin{subfigure}{.15\textwidth}
      \centering
      \includegraphics[width=.34\linewidth]{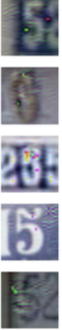}
      \caption{VAT~\cite{miyato2018virtual}}
      \label{fig:c}
    \end{subfigure}%
    \begin{subfigure}{.28\textwidth}
      \centering
      \includegraphics[width=.95\linewidth]{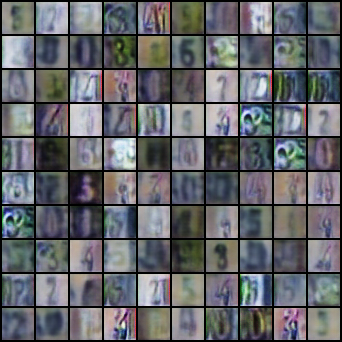}
      \caption{U-GAN (SVHN)}
      \label{fig:d}
    \end{subfigure}%
    
    \begin{subfigure}{.28\textwidth}
      \centering
      \includegraphics[width=.95\linewidth]{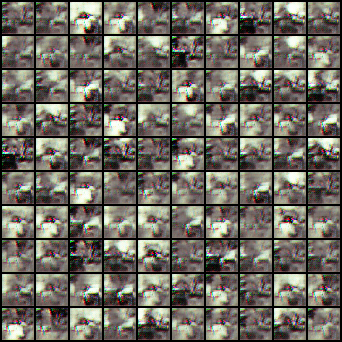}
      \caption{FM~ (CIFAR-10)}
      \label{fig:e}
    \end{subfigure}%
    \begin{subfigure}{.28\textwidth}
      \centering
      \includegraphics[width=.95\linewidth]{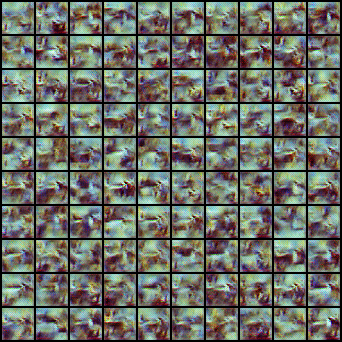}
      \caption{C-GAN~ (CIFAR-10)}
      \label{fig:f}
    \end{subfigure}%
    \begin{subfigure}{.15\textwidth}
      \centering
      \includegraphics[width=.34\linewidth]{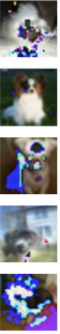}
      \caption{VAT \cite{miyato2018virtual}}
      \label{fig:g}
    \end{subfigure}%
    \begin{subfigure}{.28\textwidth}
      \centering
      \includegraphics[width=.95\linewidth]{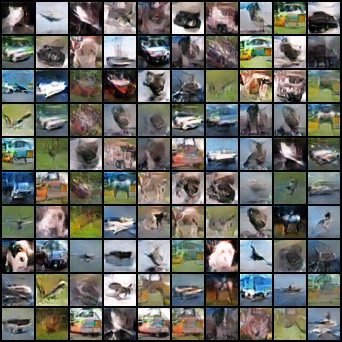}
      \caption{U-GAN (CIFAR-10)}
      \label{fig:h}
    \end{subfigure}%
    \caption{Comparison of data generated by U-GAN with the baseline algorithms. The samples generated by the baseline methods are based on our runs without any unlabeled data, and leads to significant degradation of the quality compared to the original papers (see Appendix~\ref{app_qualitative}). U-GAN generates more realistic samples compared to the rest. The VAT samples correspond to the best hyperparameter configuration, and the generated noise that has been added to the original image can be easily spotted.}
    \label{fig:gen_data}
\end{figure*}

\begin{table*}
    \centering
    \begin{small}
    \caption{Diversity and quality of the Generated data for different methods.}
    \tabcolsep=0.4cm
    \begin{tabular}{l|ll|ll}
         \hline
         \multirow{2}{*}{Method}& \multicolumn{2}{c}{Generator Entropy} & \multicolumn{2}{c}{Generator FID} \\
         \cline{2-5}
         & SVHN & CIFAR-$10$ & SVHN & CIFAR-$10$ \\
         \hline
         FM~\cite{salimans2016improved} & $2.82 \pm 0.37$ & $2.88 \pm 0.16$ & $134.1 \pm 0.8$ & $124.9 \pm 4.3$ \\
         C-GAN~\cite{dai2017good} & $3.03 \pm 0.05$ & $3.07 \pm 0.07$ & $135.2 \pm 0.7$ & $126.8 \pm 4.6$ \\
         U-GAN & $3.15 \pm 0.16$ & $3.13 \pm 0.11$ & $130.9 \pm 1.2$ & $120.6 \pm 2.3$ \\
         U-GAN + PT/VI  & $\mathbf{3.23 \pm 0.04}$ & $\mathbf{3.19 \pm 0.10}$ & $\mathbf{124.1 \pm 1.7}$ & $\mathbf{116.5 \pm 2.6}$ \\
         \hline
    \end{tabular}
    \label{tab:entropy}
    \end{small}
\end{table*}


\noindent \textbf{Generated Data (Quality)}: Next, we compare the quality of the data generated by U-GAN with that of FM, C-GAN, and VAT. Figure~\ref{fig:gen_data} provides a random set of images generated by U-GAN and the benchmark algorithms. As seen from  Figure~\ref{fig:gen_data}, U-GAN provides more realistic images for both SVHN and CIFAR-10 datasets, while both FM and C-GAN perform poorly in the absence of unlabeled data. Appendix~\ref{app_qualitative} provides a similar qualitative comparison of the generated data with FM and C-GAN when additional unlabeled data are provided, which further confirms the qualitative improvement in generated samples by U-GAN compared to the baseline algorithms. 

\noindent \textbf{Generated Data (Diversity)} One of the main challenges of training a GAN game is the mode collapse. While C-GAN~\cite{dai2017good} aims to avoid mode collapse by including an entropy term into the generator cost function, the universum loss of U-GAN implicitly regularizes the model to increase the entropy of the generated data, which in turn avoids mode collapse (also discussed in Section \ref{sec_UGAN}). Table~\ref{tab:entropy} demonstrates the advantage of U-GAN compared to the benchmark algorithms by providing the mean $\pm$ standard deviation of the class entropy and FID scores of the generated data over 10 experiment runs. As seen from this table, U-GAN outperforms FM and C-GAN in terms of the entropy of the generated samples and FID scores. Such implicit mechanism is hugely desirable for 2-player games for avoiding mode collapse. In fact, adding explicit entropy terms in U-GAN+PT/VI further improved the generator entropy. Note that, for $10$-class problems the maximum generator entropy is $\log_2 10=3.32$.  To summarize, adopting U-GAN leads to the generated samples that are more diverse and have higher quality, while maintaining desirable discriminator generalization. Additional results on MNIST data is also available in Appendix~\ref{app_UGAN_entropy}. \begin{wraptable}[20]{r}{0.45\textwidth}
	\begin{minipage}{\linewidth}
	\centering
    \caption{U-GAN vs. E-GAN.}
    \resizebox{\columnwidth}{!}{
    \begin{tabular}{lcc}
         \hline
         Method & SVHN & CIFAR-$10$ \\
         \hline
         U-GAN & $84.96 \pm 1.17$ & $68.24 \pm 1.05$ \\
         E-GAN & $\mathbf{87.42 \pm 0.53}$ & $\mathbf{70.46 \pm 0.62}$ \\
         \hline
    \end{tabular}
    }
    \label{tab:evolve} 
	\end{minipage}\hfill \\ 
	\begin{minipage}{\linewidth}
		\centering
		\includegraphics[width=65mm]{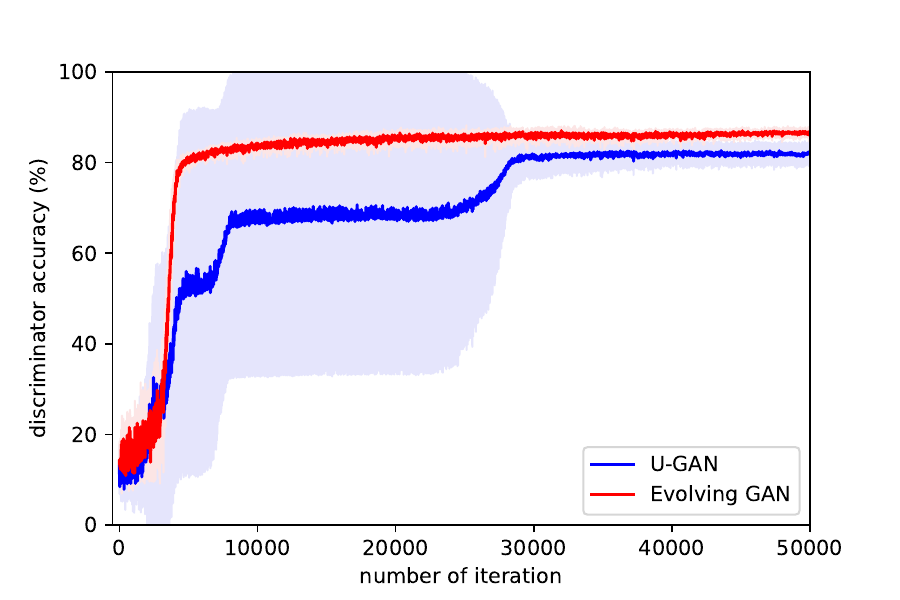}
		\captionof{figure}{Discriminator performance of U-GAN vs. E-GAN at different iterations for the SVHN dataset.}
    \label{fig:evolution}
	\end{minipage}
\end{wraptable}

\subsection{Effectiveness of Evolving GAN (E-GAN)}
\label{sec:exp_evolving_gan} 
Next, we illustrate the effectiveness of E-GAN over U-GAN. To this end, we compare the discriminator accuracy of U-GAN vs. E-GAN over 10 experimental runs for the SVHN and CIFAR-10 datasets in  Table~\ref{tab:evolve}. Table~\ref{tab:evolve} illustrates that by evolving the discriminator loss from universum to semi-supervised setting, we can better account for the changing generator samples' distribution and achieve performance gains in the discriminator accuracy. We also analyze the typical training convergence curves for SVHN dataset in Fig.~\ref{fig:evolution}. As seen in Fig.~\ref{fig:evolution}, E-GAN converges to a reasonable performance accuracy much faster. Secondly, the variation of the results over multiple runs is much smaller for E-GAN compared to U-GAN. In essence, the E-GAN provides stable and improved convergence over U-GAN. Similar results can also be seen for the CIFAR-10 (results provided in Appendix \ref{conv_curve}). The complete set of model parameters for all experiments is provided in Appendix~\ref{model_params} for reproducibility.

\section{Conclusion} \label{sec_conc}
This paper proposes to use the universum learning setting for training the discriminator in a GAN game. The proposed U-GAN game is theoretically consistent and generates more diverse and high quality data compared to baseline FM-GAN \cite{salimans2016improved} or C-GAN \cite{dai2017good} methods, while simultaneously improving the discriminator generalization. We further motivate to evolve the discriminator loss from universum to semi-supervised setting to account for the changing generator sample distribution and propose the evolving GAN (E-GAN) algorithm. The proposed E-GAN provides stable and improved convergence compared to U-GAN and further improves the  discriminator accuracy. Finally, we discuss the limitations and future research directions (moved to Appendix \ref{sec:future_research} due to space constraints).


{\small
\bibliographystyle{ieee_fullname}
\bibliography{EvolvingGANS}

\begin{thebibliography}{10}\itemsep=-1pt

\bibitem{abadi2016tensorflow}
Martin Abadi, Paul Barham, Jianmin Chen, Zhifeng Chen, Andy Davis, Jeffrey
  Dean, Matthieu Devin, Sanjay Ghemawat, Geoffrey Irving, Michael Isard, et~al.
\newblock Tensorflow: A system for large-scale machine learning.
\newblock In {\em 12th $\{$USENIX$\}$ symposium on operating systems design and
  implementation ($\{$OSDI$\}$ 16)}, pages 265--283, 2016.

\bibitem{balamurugan2013large}
P Balamurugan, Shirish Shevade, and Sundararajan Sellamanickam.
\newblock Large margin semi-supervised structured output learning.
\newblock {\em arXiv preprint arXiv:1311.2139}, 2013.

\bibitem{chapelle2009semi}
Olivier Chapelle, Bernhard Scholkopf, and Alexander Zien.
\newblock Semi-supervised learning (chapelle, o. et al., eds.; 2006)[book
  reviews].
\newblock {\em IEEE Transactions on Neural Networks}, 20(3):542--542, 2009.

\bibitem{collobert2006large}
Ronan Collobert, Fabian Sinz, Jason Weston, and L{\'e}on Bottou.
\newblock Large scale transductive svms.
\newblock {\em Journal of Machine Learning Research}, 7(Aug):1687--1712, 2006.

\bibitem{crammer02}
Koby Crammer and Yoram Singer.
\newblock On the learnability and design of output codes for multiclass
  problems.
\newblock {\em Machine learning}, 47(2-3):201--233, 2002.

\bibitem{dai2017good}
Zihang Dai, Zhilin Yang, Fan Yang, William~W Cohen, and Russ~R Salakhutdinov.
\newblock Good semi-supervised learning that requires a bad gan.
\newblock In {\em Advances in neural information processing systems}, pages
  6510--6520, 2017.

\bibitem{daniely2012multiclass}
Amit Daniely, Sivan Sabato, and Shai~Shalev Shwartz.
\newblock Multiclass learning approaches: A theoretical comparison with
  implications.
\newblock {\em arXiv preprint arXiv:1205.6432}, 2012.

\bibitem{dhar2019}
Sauptik Dhar, Vladimir Cherkassky, and Mohak Shah.
\newblock Multiclass learning from contradictions.
\newblock In {\em Advances in Neural Information Processing Systems}, pages
  8400--8410, 2019.

\bibitem{dhar2021doc3}
Sauptik Dhar and Bernardo~Gonzalez Torres.
\newblock Doc3-deep one class classification using contradictions.
\newblock {\em arXiv preprint arXiv:2105.07636}, 2021.

\bibitem{elezi2018transductive}
Ismail Elezi, Alessandro Torcinovich, Sebastiano Vascon, and Marcello Pelillo.
\newblock Transductive label augmentation for improved deep network learning.
\newblock In {\em 2018 24th International Conference on Pattern Recognition
  (ICPR)}, pages 1432--1437. IEEE, 2018.

\bibitem{feurer2019auto}
Matthias Feurer, Aaron Klein, Katharina Eggensperger, Jost~Tobias Springenberg,
  Manuel Blum, and Frank Hutter.
\newblock Auto-sklearn: efficient and robust automated machine learning.
\newblock In {\em Automated Machine Learning}, pages 113--134. Springer, Cham,
  2019.

\bibitem{kavalerov2021multi}
Ilya Kavalerov, Wojciech Czaja, and Rama Chellappa.
\newblock A multi-class hinge loss for conditional gans.
\newblock In {\em Proceedings of the IEEE/CVF Winter Conference on Applications
  of Computer Vision}, pages 1290--1299, 2021.

\bibitem{kingma2014adam}
Diederik~P Kingma and Jimmy Ba.
\newblock Adam: A method for stochastic optimization.
\newblock {\em arXiv preprint arXiv:1412.6980}, 2014.

\bibitem{lecun1998gradient}
Yann LeCun, L{\'e}on Bottou, Yoshua Bengio, and Patrick Haffner.
\newblock Gradient-based learning applied to document recognition.
\newblock {\em Proceedings of the IEEE}, 86(11):2278--2324, 1998.

\bibitem{lecun-mnisthandwrittendigit-2010}
Yann LeCun and Corinna Cortes.
\newblock {MNIST} handwritten digit database.
\newblock 2010.

\bibitem{li2017triple}
Chongxuan Li, Kun Xu, Jun Zhu, and Bo Zhang.
\newblock Triple generative adversarial nets.
\newblock {\em arXiv preprint arXiv:1703.02291}, 2017.

\bibitem{liu2019auptimizer}
Jiayi Liu, Samarth Tripathi, Unmesh Kurup, and Mohak Shah.
\newblock Auptimizer -- an extensible, open-source framework for hyperparameter
  tuning, 2019.

\bibitem{maaloe2016auxiliary}
Lars Maal{\o}e, Casper~Kaae S{\o}nderby, S{\o}ren~Kaae S{\o}nderby, and Ole
  Winther.
\newblock Auxiliary deep generative models.
\newblock In {\em International conference on machine learning}, pages
  1445--1453. PMLR, 2016.

\bibitem{miyato2018virtual}
Takeru Miyato, Shin-ichi Maeda, Masanori Koyama, and Shin Ishii.
\newblock Virtual adversarial training: a regularization method for supervised
  and semi-supervised learning.
\newblock {\em IEEE transactions on pattern analysis and machine intelligence},
  41(8):1979--1993, 2018.

\bibitem{ouali2020overview}
Yassine Ouali, C{\'e}line Hudelot, and Myriam Tami.
\newblock An overview of deep semi-supervised learning.
\newblock {\em arXiv preprint arXiv:2006.05278}, 2020.

\bibitem{pytorch}
Adam Paszke, Sam Gross, Francisco Massa, Adam Lerer, James Bradbury, Gregory
  Chanan, Trevor Killeen, Zeming Lin, Natalia Gimelshein, Luca Antiga, Alban
  Desmaison, Andreas Kopf, Edward Yang, Zachary DeVito, Martin Raison, Alykhan
  Tejani, Sasank Chilamkurthy, Benoit Steiner, Lu Fang, Junjie Bai, and Soumith
  Chintala.
\newblock Pytorch: An imperative style, high-performance deep learning library.
\newblock In H. Wallach, H. Larochelle, A. Beygelzimer, F. d Alcheuc, E. Fox,
  and R. Garnett, editors, {\em Advances in Neural Information Processing
  Systems 32}, pages 8026--8037. Curran Associates, Inc., 2019.

\bibitem{pedregosa2011scikit}
Fabian Pedregosa, Ga{\"e}l Varoquaux, Alexandre Gramfort, Vincent Michel,
  Bertrand Thirion, Olivier Grisel, Mathieu Blondel, Peter Prettenhofer, Ron
  Weiss, Vincent Dubourg, et~al.
\newblock Scikit-learn: Machine learning in python.
\newblock {\em the Journal of machine Learning research}, 12:2825--2830, 2011.

\bibitem{salimans2016improved}
Tim Salimans, Ian Goodfellow, Wojciech Zaremba, Vicki Cheung, Alec Radford, and
  Xi Chen.
\newblock Improved techniques for training gans.
\newblock In {\em Advances in neural information processing systems}, pages
  2234--2242, 2016.

\bibitem{selvaraj2012extension}
Sathiya~Keerthi Selvaraj, Sundararajan Sellamanickam, and Shirish Shevade.
\newblock Extension of tsvm to multi-class and hierarchical text classification
  problems with general losses.
\newblock {\em arXiv preprint arXiv:1211.0210}, 2012.

\bibitem{shi2018transductive}
Weiwei Shi, Yihong Gong, Chris Ding, Zhiheng~MaXiaoyu Tao, and Nanning Zheng.
\newblock Transductive semi-supervised deep learning using min-max features.
\newblock In {\em Proceedings of the European Conference on Computer Vision
  (ECCV)}, pages 299--315, 2018.

\bibitem{sinz2007priori}
FH Sinz.
\newblock {\em A priori knowledge from non-examples}.
\newblock PhD thesis, Eberhard-Karls-Universit{\"a}t T{\"u}bingen, Germany,
  2007.

\bibitem{sinz08}
FH. Sinz, O. Chapelle, A. Agarwal, and B. Sch{\"o}lkopf.
\newblock An analysis of inference with the universum.
\newblock In {\em Advances in neural information processing systems 20}, pages
  1369--1376, NY, USA, Sept. 2008. Curran.

\bibitem{vapnik06}
V. Vapnik.
\newblock {\em {Estimation of Dependences Based on Empirical Data (Information
  Science and Statistics)}}.
\newblock Springer, Mar. 2006.

\bibitem{vargas2019robustness}
Danilo~Vasconcellos Vargas and Shashank Kotyan.
\newblock Robustness assessment for adversarial machine learning: Problems,
  solutions and a survey of current neural networks and defenses.
\newblock {\em arXiv preprint arXiv:1906.06026}, 2019.

\bibitem{wang2013dynamic}
Bo Wang, Zhuowen Tu, and John~K Tsotsos.
\newblock Dynamic label propagation for semi-supervised multi-class multi-label
  classification.
\newblock In {\em Proceedings of the IEEE international conference on computer
  vision}, pages 425--432, 2013.

\bibitem{zhang2017universum}
Xiang Zhang and Yann LeCun.
\newblock Universum prescription: Regularization using unlabeled data.
\newblock In {\em Proceedings of the AAAI Conference on Artificial
  Intelligence}, page 2907–2913, 2017.

\bibitem{zien2007transductive}
Alexander Zien, Ulf Brefeld, and Tobias Scheffer.
\newblock Transductive support vector machines for structured variables.
\newblock In {\em Proceedings of the 24th international conference on Machine
  learning}, pages 1183--1190, 2007.

\end{thebibliography}
}

\clearpage
\appendix

\section{Proofs}
\subsection{Proof of Proposition \ref{prop_max_contradiction} }
See Proposition 1 in \cite{dhar2019}.

\subsection{Proof of Proposition \ref{prop_tran_usvm} }
See Proposition 3 in \cite{dhar2019}.

\subsection{Proof of Proposition \ref{prop_tran_tsvm} } \label{sec_proof_unified}
The proof follows from analyzing the contribution of each sample to the loss function. The contribution of labeled samples are the same in both problems as they are defined identically. For unlabeled data, when $\epsilon = 1$ and $\psi_{\epsilon}(x) = \text{min}(x,\epsilon) = \text{min}(x,1)$ we have
\begin{align} 
& \underset{\mathbf{w}_1 \ldots \mathbf{w}_L ,\boldsymbol\xi}{\text{min}} \sum\limits_{i=1}^n \xi_{i} +  C_U \sum\limits_{i=n+1}^{n+mL} \xi_{i}  && \nonumber \\
&\text{s.t.} \quad \xi_i = \underset{k \in \mathcal{Y}}{\text{max}} \; \{ 1-\delta_{ik} + \mathbf{w}_k^T\mathbf{x}_i - \mathbf{w}_{y_i}^T\mathbf{x}_i \} ; \; i=1 \ldots n && \nonumber \\
&\xi_{i} = \text{min}\{\underset{k \in \mathcal{Y}}{\text{max}} \; \{ \epsilon (1-\delta_{i^{\prime}k}) + \mathbf{w}_k^T\mathbf{x}_{i} - \mathbf{w}_{y_{i}}^T\mathbf{x}_{i} \}, 1\} && \nonumber \\
& \quad \quad \quad \quad \quad \quad \quad \quad \quad \quad \quad \quad \quad \forall i = n+1 \ldots n+mL && \nonumber
\end{align}
\noindent Let the final solution be $\mathbf{w} = [\mathbf{w}_1,\ldots \mathbf{w}_L]$. We analyze the contribution of the unlabeled samples $(\mathbf{x}_{i},y_{i})_{i = n+1}^{n+mL} $ i.e. $( (\mathbf{x}_{i^{\prime}}^*,y_{i^{\prime}l}^*)_{l = 1}^{L})_{i^{\prime} = 1}^{m} $. Each $\mathbf{x}_{i^\prime}^*$ is introduced multiple times with labels $y_{i^{\prime}1}=1 \ldots y_{i^{\prime}L}=L$ through the transformation in Definition \eqref{def_transformation}. WLOG we assume with the final solution for the samples $(\mathbf{x}_{i^\prime}^*, y=1) \ldots (\mathbf{x}_{i^\prime}^*, y=L)$ we have,
\begin{align}
\mathbf{w}_k^{\top} \mathbf{x}_{i^\prime}^* >  \mathbf{w}_l^{\top} \mathbf{x}_{i^\prime}^* ; \quad \forall l \neq k \quad \text{i.e. } y_{i^{\prime}}^* = k && \nonumber 
\end{align}
The above condition means that $\{\zeta_{i^{\prime} \ell}\}_{\ell=1}^L$ in~\eqref{eq_cs_transductive_loss} can be obtained as follows:
\begin{align}
&\zeta_{i^{\prime} 1} = 1&&\nonumber \\ 
& \quad \vdots &&\nonumber\\
&\zeta_{i^{\prime} k} = \underset{l\neq k}{\text{max}}[1+ \mathbf{w}_l^{\top} \mathbf{x}_{i^{\prime}}^* - \mathbf{w}_k^{\top} \mathbf{x}_{i^{\prime}}^* ]&&\nonumber\\
&\quad \quad  \vdots &&\nonumber\\
&\zeta_{i^{\prime} L} = 1 &&\nonumber
\end{align}
Hence, the overall contribution of unlabeled data becomes,
\[ \underset{l}{\text{max}}[1+ \mathbf{w}_l^{\top}\mathbf{x}_{i^{\prime}}^* - \mathbf{w}_{y_{i^{\prime}}^*}^{\top}\mathbf{x}_{i^{\prime}}^*] + (L-1) \] That is sum of the constraint in \eqref{eq_cs_transductive_loss} and a constant. Hence the solution to the \eqref{eq_cs_unified_loss} also solves \eqref{eq_cs_transductive_loss} \qed

\subsection{Proof of Proposition \ref{prop_UGAN} } \label{sec_proof_UGANprop}
\noindent We divide the proof of this proposition into two parts as shown in Lemmas \eqref{lemma_best_disc} and \eqref{lemma_best_gen}
\begin{lemma} \label{lemma_best_disc}
Under assumption \ref{assumption_bayes} for a fixed $G$, the optimal discriminator $D^*$ that solves \eqref{eq_gan_play1} with any $C_G\leq \frac{\mathbbm{P}_{\mathcal{X}}(\mathbf{x} \in \Omega)}{\mathbbm{P}_z(\mathbf{\hat{x}} \in \Omega)}$; $\mathbf{\hat{x}} = G(z)$ and $\Omega = \mathcal{X} \cap G(z)$ satisfies,
\begin{align} \label{eq_optimal_Disc1}
&h_{D^*(\mathbf{x})} \quad = \quad h^*(\mathbf{x}); \quad \forall \mathbf{x} \in \mathcal{X} && \nonumber\\
&\quad \quad \quad \; \; \neq \quad \{1,\ldots, L\} ; \quad \forall \mathbf{x} \notin \mathcal{X} && 
\end{align}
This implies, For the decision rule in \eqref{eq_dec_rule} with linear parameterization $f_l = \mathbf{w}_l^{\top} \mathbf{x} ; \; \forall l \in \mathcal{Y} = \{1 \ldots L \} $ we have,
\begin{flalign} \label{eq_optimal_Disc2}
D^*(\mathbf{x})  \;\Rightarrow \; \left\{
\begin{array}{l l}
    \mathbf{w}_y^{\top}\mathbf{x} - \underset{k \neq y }{\text{max}}\; \mathbf{w}_k^{\top}\mathbf{x} >0 \quad  \text{if} \; (\mathbf{x},y) \sim \mathcal{D}_{\mathcal{X}} \times \mathcal{D}_{\mathcal{Y}}  \\
    \mathbf{w}_k^{\top} \mathbf{x} = \mathbf{w}_l^{\top} \mathbf{x}; \quad \forall k,l \in \{1,\ldots,L\} \quad  \text{else}
\end{array}\right.
\end{flalign}
\end{lemma}
\noindent \textbf{Proof} The proof follows by partitioning the error probabilities in \eqref{eq_gan_play1} into different event spaces. We define, $A = \mathcal{X} - G(z); \; B = G(z) - \mathcal{X}; \; \Omega = \mathcal{X} \cap G(z)$. \\
\begin{figure}[h]
\centering
\includegraphics[width=6.5cm]{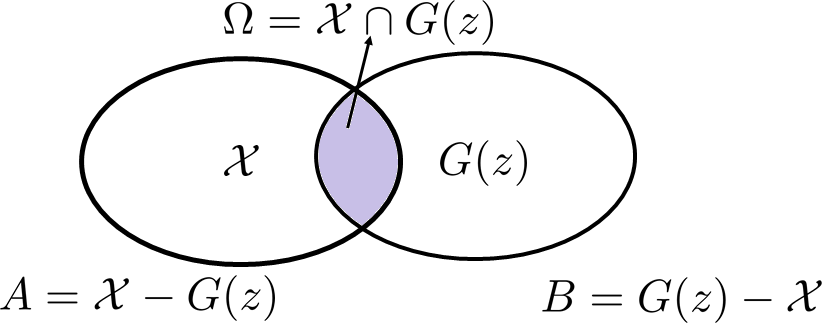}
\caption{Partitioning the event space} \label{fig:sampleSpace} 
\end{figure} 
Next, we rewrite,
\begin{align}
& L_D =  \underset{XY}{\mathbbm{E}}[\mathbbm{1}_{y=h_{D(\mathbf{x})}}] + C_G\; \underset{z}{\mathbbm{E}}[\mathbbm{1}_{h_{D(G(z))} \notin \mathcal{Y}}] && \nonumber \\ 
& =  \underset{Y|X}{\mathbbm{E}}[\mathbbm{1}_{y=h_{D(\mathbf{x})}}|\mathbf{x} \in A] \underset{X}{\mathbbm{P}}(\mathbf{x} \in A) \quad \text{(Total probability)}&& \nonumber \\ 
& \quad +  \underset{Y|X}{\mathbbm{E}}[\mathbbm{1}_{y=h_{D(\mathbf{x})}}|\mathbf{x} \in \Omega] \underset{X}{\mathbbm{P}}(\mathbf{x} \in \Omega) && \nonumber \\
&\quad + \; C_G \underset{z}{\mathbbm{E}} [\mathbbm{1}_{\underset{k  \in \mathcal{Y}}{\bigcap} h_{D(G(z))} \neq k}|G(z) \in B] \; {\mathbbm{P}}(G(z) \in B) &&\nonumber \\
&\quad + \;C_G \underset{z}{\mathbbm{E}} [\mathbbm{1}_{\underset{k  \in \mathcal{Y}}{\bigcap} h_{D(G(z))} \neq k}|G(z) \in \Omega] \; {\mathbbm{P}}(G(z) \in \Omega)  && \nonumber \\
& = \tcircle{\small{a}} + \tcircle{\small{b}} + \tcircle{\small{c}} && \label{eq_partition}
\end{align}
where,

\begin{align} \label{eq_abc}
&\tcircle{\small{a}} = \underset{Y|X}{\mathbbm{E}}[\mathbbm{1}_{y=h_{D(\mathbf{x})}}|\mathbf{x} \in A] \underset{X}{\mathbbm{P}}(\mathbf{x} \in A) &&  \\ 
&\tcircle{\small{b}} = \underset{Y|X}{\mathbbm{E}}[\mathbbm{1}_{y=h_{D(\mathbf{x})}}|\mathbf{x} \in \Omega] \underset{X}{\mathbbm{P}}(\mathbf{x} \in \Omega) && \nonumber \\
&\quad + \; C_G \underset{z}{\mathbbm{E}} [\mathbbm{1}_{\underset{k  \in \mathcal{Y}}{\bigcap} h_{D(G(z))} \neq k}|G(z) \in B] \; {\mathbbm{P}}(G(z) \in B) &&\nonumber \\
&\tcircle{\small{c}} = \;C_G \underset{z}{\mathbbm{E}} [\mathbbm{1}_{\underset{k  \in \mathcal{Y}}{\bigcap} h_{D(G(z))} \neq k}|G(z) \in \Omega] \; {\mathbbm{P}}(G(z) \in \Omega)  && \nonumber 
\end{align}

\noindent Note that for the decision rule in \eqref{eq_dec_rule}, we have,
\begin{align}
&\mathbbm{1}_{y=h_{D(\mathbf{x})}} = \mathbbm{1}_{ \mathbf{w}_y^{\top} \mathbf{x} - \underset{k \neq y}{\text{max}} \; \mathbf{w}_k^{\top} \mathbf{x} >0} && \nonumber \\
&\text{and from Proposition \eqref{prop_max_contradiction} we have,} &&\nonumber \\
&\mathbbm{1}_{\underset{k  \in \mathcal{Y}}{\bigcap} h_{D(G(z))} \neq k} = \mathbbm{1}_{|(\mathbf{w}_k^\top \mathbf{x}^{*}-\underset{l=1 \ldots L}{\text{max}}\mathbf{w}_l^\top \mathbf{x}^{*})| = 0 ; \; \forall k \in \{1,\ldots, L\}} && \nonumber\\
&\quad \quad \quad \quad \quad \quad  = \mathbbm{1}_{\mathbf{w}_k^\top \mathbf{x}^{*} = \mathbf{w}_l^\top \mathbf{x}^{*} ; \quad \forall (k,l) \in \mathcal{Y}} && \nonumber
\end{align}

\noindent Hence, \eqref{eq_abc}  translates to, 
\begin{align}  \label{eq_abc2}
&\tcircle{\small{a}} = \underset{Y|X}{\mathbbm{E}}[\mathbbm{1}_{ \mathbf{w}_y^{\top} \mathbf{x} - \underset{k \neq y}{\text{max}} \; \mathbf{w}_k^{\top} \mathbf{x} >0} |\mathbf{x} \in A] \; \underset{X}{\mathbbm{P}}(\mathbf{x} \in A)  &&   \\ 
&\tcircle{\small{b}} = \underset{Y|X}{\mathbbm{E}}[\mathbbm{1}_{\mathbf{w}_y^{\top} \mathbf{x} - \underset{k \neq y}{\text{max}} \; \mathbf{w}_k^{\top} \mathbf{x} >0} | \mathbf{x} \in \Omega] \underset{X}{\mathbbm{P}}(\mathbf{x} \in \Omega) &&  \nonumber  \\
& \quad \quad \quad  + C_G \underset{z}{\mathbbm{E}} [\mathbbm{1}_{\mathbf{w}_k^{\top} G(z) = \mathbf{w}_l^{\top} G(z); \;  \forall l,k \in \mathcal{Y}}|G(z) \in \Omega] \; {\mathbbm{P}_z}(G(z) \in \Omega)   \nonumber && \\
&\tcircle{\small{c}} = C_G \; \underset{z}{\mathbbm{E}} [\mathbbm{1}_{\mathbf{w}_k^{\top} G(z) = \mathbf{w}_l^{\top} G(z); \;  \forall l,k \in \mathcal{Y}}|G(z) \in B] \; {\mathbbm{P}_z}(G(z) \in B)    \nonumber  &&
\end{align}

\noindent Under assumption \eqref{assumption_bayes}, the overall loss $L_D$ (in \eqref{eq_partition}) is maximized if $D^*(x)$ follows \eqref{eq_optimal_Disc2}. Why? Note that for such a $D^*$, \eqref{eq_partition} translates to,
\begin{align}
&\tcircle{\small{a}} = \underset{X}{\mathbbm{P}}(\mathbf{x} \in A) \quad (\text{max. possible value})&& \nonumber \\
&\tcircle{\small{b}} =  \underset{X}{\mathbbm{P}}(\mathbf{x} \in \Omega) \quad (\text{max. possible value}) &&  \nonumber
\end{align}
\noindent Since, $\mathbbm{1}_{ \mathbf{w}_y^{\top} \mathbf{x} - \underset{k \neq y}{\text{max}}\; \mathbf{w}_k^{\top} \mathbf{x} >0}$ and $\mathbbm{1}_{\mathbf{w}_k^\top \mathbf{x} = \mathbf{w}_l^\top \mathbf{x} ; \quad \forall (k,l) \in \mathcal{Y}}$ are mutually exclusive; only one event is triggered. For $C_G\leq \frac{\mathbbm{P}_{\mathcal{X}}(\mathbf{x} \in \Omega)}{\mathbbm{P}_z(\mathbf{\hat{x}} \in \Omega)}$, the first term dominates and maximizes $L_D$. Finally,
\begin{align}
&\tcircle{\small{c}} =  \underset{z}{\mathbbm{P}}(G(z) \in B) \quad (\text{max. possible value}) &&  \nonumber
\end{align}

\noindent This justifies setting $D^*$ as in \eqref{eq_optimal_Disc2} to maximize $L_D$. It is straightforward to see \eqref{eq_optimal_Disc2} $\Rightarrow$ \eqref{eq_optimal_Disc1}, under the above parameterization.
\qed

\begin{lemma} \label{lemma_best_gen}
For the fixed $D^*$ and $C_G$ in Lemma \eqref{lemma_best_disc}, the optimal $G^*$ that maximizes \eqref{eq_gan_play2} ensures $G(z) \subseteq \mathcal{X} ; \; \forall z \sim P_z$ i.e. support of $P_G^*$ is contained in $\mathcal{X}$. 
\end{lemma}

\noindent \textbf{Proof} 
\begin{align}
&L_G = \underset{z}{\mathbbm{E}} [\mathbbm{1}_{h_{D^*(G(z)) } \in  \mathcal{Y}} | G(z) \in \Omega] \; {\mathbbm{P}}(G(z) \in \Omega) &&\nonumber \\
&\quad \quad +\;  \underset{z}{\mathbbm{E}} [\mathbbm{1}_{h_{D^*(G(z))} \in  \mathcal{Y}} | G(z) \in B] \; {\mathbbm{P}}(G(z) \in B) && \nonumber
\end{align}
From \eqref{eq_optimal_Disc1}, $\underset{z}{\mathbbm{E}} [\mathbbm{1}_{h_{D^*(G(z))} \in  \mathcal{Y}} | G(z) \in B] = 0$. Hence, $L_G$ is maximized if ${\mathbbm{P}}(G(z) \in \Omega) = 1$.
\qed

\noindent Finally combining Lemma \eqref{lemma_best_disc} and \eqref{lemma_best_gen} we get Proposition \eqref{prop_UGAN} \qed

\subsection{Proof of Claim \ref{claim_UGAN} } \label{sec_proof_UGANclaim}
The proof follows by analyzing the weightage of the terms in \eqref{eq_partition} \textcircled{\small{b}}. For the case without the Assumption \eqref{assumption_bayes}, all that we need is to select a, \[C_G \leq \frac{\underset{Y|X}{\mathbbm{E}}[\mathbbm{1}_{\mathbf{w}_y^{\top} \mathbf{x} - \underset{k \neq y}{\text{max}} \; \mathbf{w}_k^{\top} \mathbf{x} >0} | \mathbf{x} \in \Omega] \; \underset{X}{\mathbbm{P}}(\mathbf{x} \in \Omega) }{\underset{z}{\mathbbm{E}} [\mathbbm{1}_{\mathbf{w}_k^{\top} G(z) = \mathbf{w}_l^{\top} G(z); \;  \forall l,k \in \mathcal{Y}}|G(z) \in \Omega] \; {\mathbbm{P}_z}(G(z) \in \Omega)}\] For, such a selected $C_G$, the proposition \eqref{prop_UGAN} holds without Assumption \eqref{assumption_bayes} \qed

\newpage 

\section{Additional Empirical results}

\subsection{Baseline comparisons for the Transductive C\&S Loss solved using Proposition \eqref{prop_tran_tsvm}} 
\label{app_TSVM_baselines}
\begin{figure}[h]
\centering
\includegraphics[width=8.5cm]{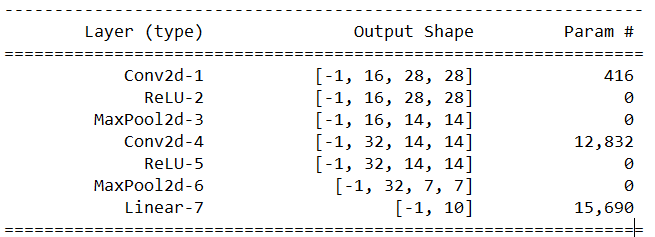}
\caption{CNN architecture summary used for MNIST example.} \label{fig:basicCNN} 
\end{figure}
In this work we solve the Transductive C\& S problem in \eqref{eq_cs_transductive_loss} using Proposition \eqref{prop_tran_tsvm} and compare its performance against traditional solvers which uses a switching algorithm \cite{zien2007transductive, selvaraj2012extension, balamurugan2013large}. Note that, both Transductive SVM (T-SVM) and semi-supervised SVM solves the same underlying optimization formulations. Our proposition \ref{prop_tran_tsvm} provides the following advantages,
\begin{enumerate}
    \item Our approach is an extension to \cite{collobert2006large} for multiclass problems, and similarly scales to large problems. In addition, we can now avoid using switching algorithms typically adopted for multiclass Transductive C \&S formulation \cite{zien2007transductive, selvaraj2012extension, balamurugan2013large}, and adds significant computational load for solving the transductive C \&S loss. 
    \item Further, now the formulation \eqref{eq_cs_transductive_loss} can be easily implemented in most popular deep learning frameworks \cite{pytorch,pedregosa2011scikit,abadi2016tensorflow} and solved through the state-of-art first order solvers supported in these frameworks.
\end{enumerate}
To validate the statistical performance of our approach we further baseline our implementation against existing T-SVM benchmarks \cite{zien2007transductive,wang2013dynamic}. Table \ref{tab_TSVM} provides the results on two datasets. Here we report the mean $\pm$ std. deviation of the test accuracies over 10 runs of the experimental setting discussed below, \\ \\
{\sc Coil Dataset \cite{chapelle2009semi}} \footnote{publicly available at \url{http://olivier.chapelle.cc/ssl-book/benchmarks.html}}: This is a 6 - class classification problem. We report the performance of the standard C\&S \eqref{eq_cs_loss} vs. Transductive C\&S \eqref{eq_cs_transductive_loss} losses over 10 random partitioning of the data. In each partition we randomly select $n = 100$ training samples (and remaining as test samples) following \cite{zien2007transductive}. For this experiment we use linear parameterization. Further,
\begin{itemize}
    \item \underline{C\&S loss} we use an Adam optimizer \cite{kingma2014adam} with, batchSize = $100$, No. of epochs = $5000$, step size = $0.005$. Further increase in epochs does not provide any improvement.
    \item \underline{Transductive C\&S} we use an Adam optimizer with, batchSize = $250$, No. of epochs = $50000$, step size = $0.005$.
\end{itemize} 
{\sc MNIST Dataset \cite{lecun1998gradient}}: This is a 10 - class classification problem. For this experiment following \cite{wang2013dynamic} we use 1\% ($n = 600$) samples as training. Here we rather use a very simple CNN architecture shown in Fig \ref{fig:basicCNN}. Further,  
\begin{itemize}
    \item \underline{C\&S loss} we use an Adam optimizer with, batchSize = $100$, No. of epochs = $20000$, step size = $0.001$. Further increase in epochs does not provide any improvement.
    \item \underline{Transductive C\&S} we use an Adam optimizer with, batchSize = $250$, No. of epochs = $25000$, step size = $0.001$.
\end{itemize} 
As seen from the results in Table \ref{tab_TSVM} the implementation through the transformation in Definition \eqref{def_transformation} and Proposition \eqref{prop_tran_tsvm} we can obtain similar statistical performance. Here, in each experiment we randomly select the training samples in the same proportion as mentioned above. We use the complete test data. The results show that solving the transductive C \&S loss using the transformation (in Definition \ref{def_transformation}) and Proposition \eqref{prop_tran_tsvm} provides similar statistical performance as the existing benchmarks.

\begin{table*}
\caption{Mean ($\pm$ standard deviation) of the test accuracies (in \%) over 10 runs of the experimental setting.} 
\label{tab_TSVM}
\centering
\begin{tabular}{cccc}  
\hline
Dataset  & \specialcell{C\&S Hinge}  & \specialcell{Transductive C\&S \\(\textbf{Ours})} & \specialcell{Transductive C\&S} \\
\hline
Coil & $73.39\pm 1.31$ & $74.32 \pm 1.11$ & $74.58$ \cite{zien2007transductive} \\
MNIST & $94.61 \pm 0.8$ & $96.7 \pm 0.48$ & $95.13$ \cite{wang2013dynamic} \\
\hline
\end{tabular}
\end{table*}

\subsection{Additional Analysis of the U-GAN formulation in  section \ref{sec_UGAN} using MNIST} 
\label{app_UGAN_entropy}

\begin{figure}[h]
    \centering
    \begin{subfigure}{.4\textwidth}
      \centering
      \includegraphics[width=\linewidth]{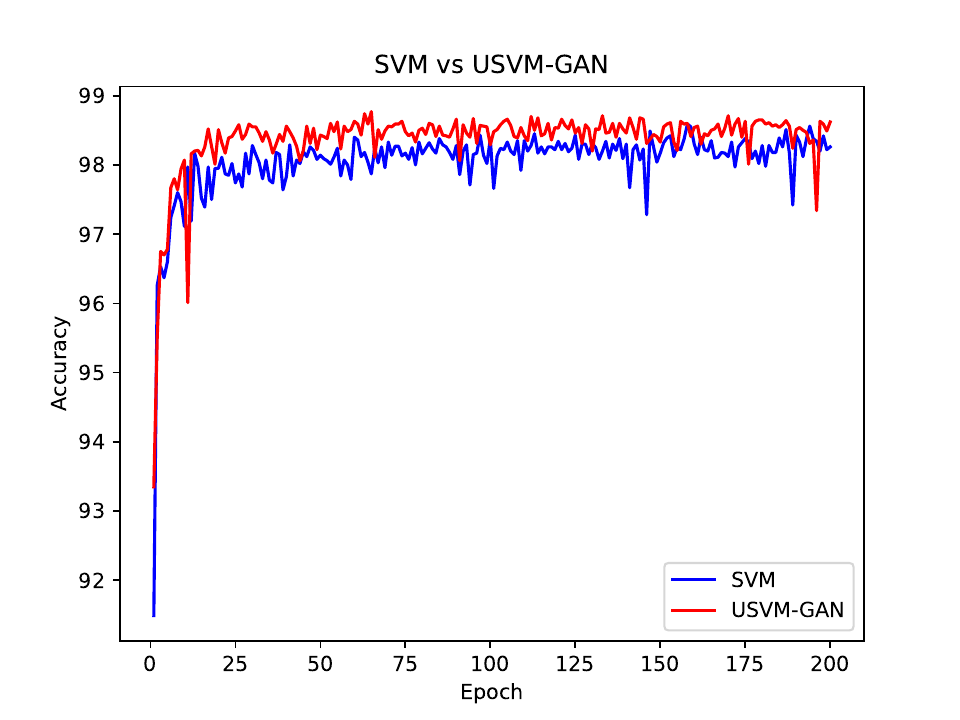}
      \caption{SVM vs. U-GAN's Discriminator performance.}
      \label{fig:usvm_gan}
    \end{subfigure}\hfill%
    \begin{subfigure}{.25\textwidth}
      \centering
      \includegraphics[width=\linewidth]{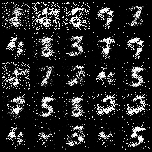}
      \caption{without $L_{\rm Disc}$}
      \label{fig:gan_sample}
    \end{subfigure}\hspace{0.2cm}%
    \begin{subfigure}{.25\textwidth}
      \centering
      \includegraphics[width=\linewidth]{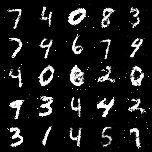}
      \caption{using $L_{\rm Disc}$}
      \label{fig:gan_sample2}
    \end{subfigure}%
    \caption{Discriminator and Generator performance comparisons for U-GAN on MNIST data.}
    \label{fig:evolution_2}
\end{figure}


This section further consolidates our U-GAN formulation in \ref{sec_UGAN}. Note that, different from previous approaches used under multiclass settings \cite{salimans2016improved} \cite{dai2017good}; here we use Universum loss for the discriminator. Further for training the generator, we combine the Feature Matching loss with a dominating surrogate of the loss in \eqref{eq_gan_play2}. Slightly different from the U-GAN (hinge) formulation in \eqref{eq_ugan_hinge}, we use the following $L_{\rm Disc}$ as the dominating surrogate, 

\begin{align}\label{eq:gen_loss}
&\min_{\boldsymbol{\theta}}\ L_{\rm FM}(\boldsymbol{\theta})+\lambda\cdot L_{\rm Disc}(\boldsymbol{\theta}) \\
&{\rm s.t.}\ \ \, L_{\rm FM}(\boldsymbol{\theta})=||\mathbb{E}[\phi(G(z;\boldsymbol{\theta}))]-\mathbb{E}[\phi(x)]||_1 , \\
&\qquad L_{\rm Disc}(\boldsymbol{\theta})=\sum_{i=1}^m  \max \log D(G(z;\boldsymbol{\theta}))\ .
\end{align} 

We use the same discriminator loss as in \eqref{eq_ugan_hinge}. For this section we refer to this formulation using the \eqref{eq:gen_loss} generator loss as U-GAN.

Our overall goal is to highlight,
\begin{itemize}
    \item Improved diversity of the U-GAN generated data (high classification labels entropy on generated data) compared to \cite{salimans2016improved}. 
    \item Effect of using the additional loss term $L_{\rm Disc}$ in the generator loss. 
\end{itemize}

Firstly, we confirm that the U-GAN discriminator achieves similar (or better) generalization compared to standard inductive learning using a traditional C\&S Hinge loss in \eqref{eq_cs_loss} (see Fig \ref{fig:usvm_gan}). For this experiment we do not see a significant improvement in discriminator generalization for U-GAN. \cite{salimans2016improved} also provides similar performance. However, similar to the results reported in Table \ref{tab:entropy}, we see significant improvement in the generated data diversity for U-GAN compared to \cite{salimans2016improved}. Here, after convergence of the GAN games we generate $1000$ samples and calculate the entropy of the classes by running the samples through the discriminator. For the U-GAN, we get entropy of $3.29$ while for $L+1$-class classifier \cite{salimans2016improved} we have $3.01$. Note that, the maximum entropy for a $10$-class setting is $\log 10=3.32$.

Next we explore the quality of the data generated by U-GAN and the effect of the additional loss component $L_{\rm Disc}$ in the generator loss. Using only the FM loss results to `salt' noise in the generated images by the GAN's generator (see Fig.~\ref{fig:gan_sample}). Rather, adding the additional $L_{\rm Disc}$ component removes this `salt' noise and provides near realistic digit data (Fig.~\ref{fig:gan_sample2}).  This shows the utility of using a good surrogate for the Generator in \eqref{eq_gan_play2} in addition to the FM loss.

\subsection{Comparison of the U-GAN generated data quality vs. state-of-the-art} 
\label{app_qualitative}

\begin{figure*}[h]
    \centering
    \begin{subfigure}{.28\textwidth}
      \centering
      \includegraphics[width=.95\linewidth]{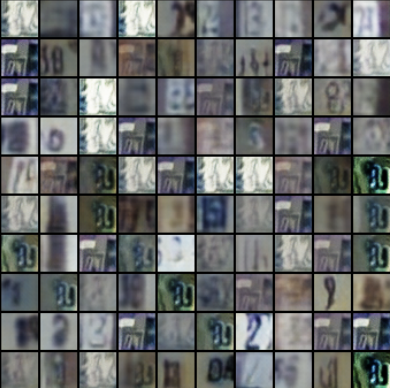}
      \caption{FM~\cite{salimans2016improved} (SVHN)}
      \label{fig:aa}
    \end{subfigure}%
    \begin{subfigure}{.28\textwidth}
      \centering
      \includegraphics[width=.95\linewidth]{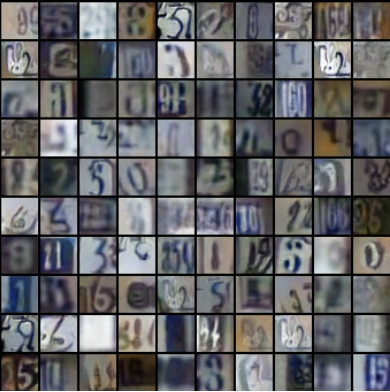}
      \caption{C-GAN~\cite{dai2017good} (SVHN)}
      \label{fig:ba}
    \end{subfigure}%
    \begin{subfigure}{.15\textwidth}
      \centering
      \includegraphics[width=.34\linewidth]{svhn-vat.png}
      \caption{VAT~\cite{miyato2018virtual}}
      \label{fig:ca}
    \end{subfigure}%
    \begin{subfigure}{.28\textwidth}
      \centering
      \includegraphics[width=.95\linewidth]{svhn-ugan.png}
      \caption{U-GAN (SVHN)}
      \label{fig:da}
    \end{subfigure}%
    
    \begin{subfigure}{.28\textwidth}
      \centering
      \includegraphics[width=.95\linewidth]{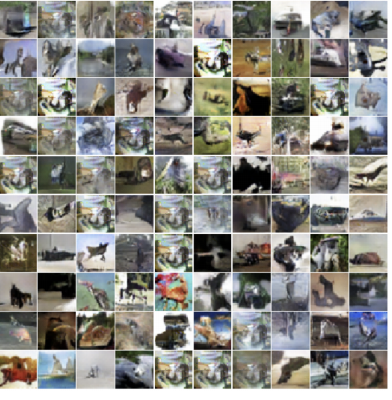}
      \caption{FM~ (CIFAR-10)}
      \label{fig:ea}
    \end{subfigure}%
    \begin{subfigure}{.28\textwidth}
      \centering
      \includegraphics[width=.95\linewidth]{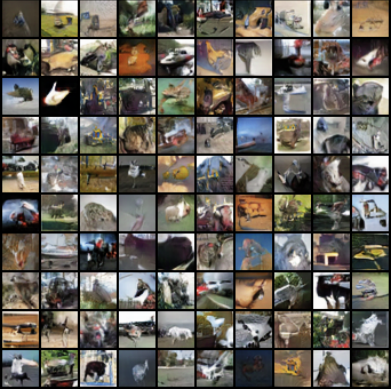}
      \caption{C-GAN~\cite{dai2017good} (CIFAR-10)}
      \label{fig:fa}
    \end{subfigure}%
    \begin{subfigure}{.15\textwidth}
      \centering
      \includegraphics[width=.34\linewidth]{cifar-vat.png}
      \caption{VAT~\cite{miyato2018virtual}}
      \label{fig:ga}
    \end{subfigure}%
    \begin{subfigure}{.28\textwidth}
      \centering
      \includegraphics[width=.95\linewidth]{cifar-ugan.png}
      \caption{U-GAN (CIFAR-10)}
      \label{fig:ha}
    \end{subfigure}%
    \caption{An example of data generated by U-GAN and state-of-the-art algorithms (Top row is the SVHN data and Bottom row is the CIFAR-10 data). The samples generated by the benchmark methods are copied from the original papers and therefore use the unlabeled data as well. Despite some low-quality, non-representative data generated by U-GAN, similar to FM it generates the most realistic samples. The VAT samples are the ones corresponding the the best-performing hyperparameter configuration, and the generated noise that has been added to the original image can be easily spotted.}
    \label{fig:gen_data_unl}
\end{figure*}

Finally we also compare the quality of the U-GAN generated data in section \ref{sec_results_UGAN_effective} of the paper with those reported for FM-GAN\cite{salimans2016improved}, C-GAN \cite{dai2017good} and VAT \cite{miyato2018virtual}. Note however, the results reported for FM-GAN\cite{salimans2016improved}, C-GAN \cite{dai2017good} leverage additional unlabeled data through  semi-supervised settings. As seen from the results in Fig \ref{fig:gen_data_unl}, U-GAN generates almost similar quality images, even without using any additional unlabeled samples. This sheds a very positive note for the proposed U-GAN approach. Since the Evolving GAN algorithm stops generator training during the semi-supervised learning phase when $\epsilon >0$ (see Algorithm \ref{algo:evolving_gan}), it provides no additional improvement on the quality of the generated data. The evolving phase (when $\epsilon >0$) mainly targets the discriminator performance at that stage.

\subsection{Convergence curve for U-GAN vs E-GAN for CIFAR-10 dataset.}  \label{conv_curve}

\begin{figure}[h]
      \centering
      \includegraphics[width=\linewidth]{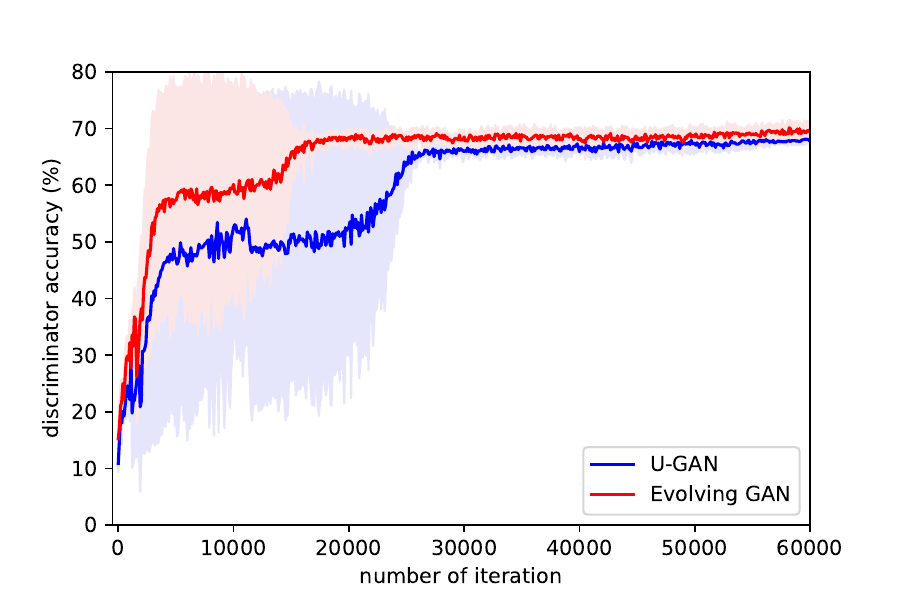}
      \label{fig:egan_cifar}
    \caption{Discriminator performance of U-GAN vs. E-GAN at different iteration of GAN training for CIFAR-10 data set.}
    \label{fig:evolution_CIFAR}
\end{figure}

Here we present the convergence curve of the U-GAN's and E-GAN's discriminator loss. As also seen for SVHN dataset in Fig.  \ref{fig:evolution}, the E-GAN convergence curve exhibit more stable and faster convergence rates. This further consolidates the need to evolve the discriminator loss from universum $\rightarrow$ semi-supervized setting.

\newpage 

\section{Reproducibility : Experiment setups, Network Architecture and Selected Model hyperparameters} 
\label{model_params}
All our experiments were performed on Amazon AWS cloud servers, using a p3.8xlarge instance with 4 NVIDIA V100 Tensor Core GPUs. We use Auptimizer ~\cite{liu2019auptimizer} library to analyze performance of different hyperparameters detailed below. 

For the sections below we use $C_U$ and $C_{gen}$ ($\hat{C}_U$ in eq.\eqref{eq_cs_unified_loss}) to represent the loss multipliers during the universum and semi-supervised setting respectively. From our analysis its clear that $C_U$ value of $0.5$ yields the ideal performance for both datasets, with the model being sensitive to it's value. Optimal $C_{gen}$ can vary for both the datasets, but the model performance remains fairly robust to it within a range of (0.1,1.0).

\subsection{SVHN data}

\begin{figure*}[h]
\centering
\includegraphics[width=\textwidth]{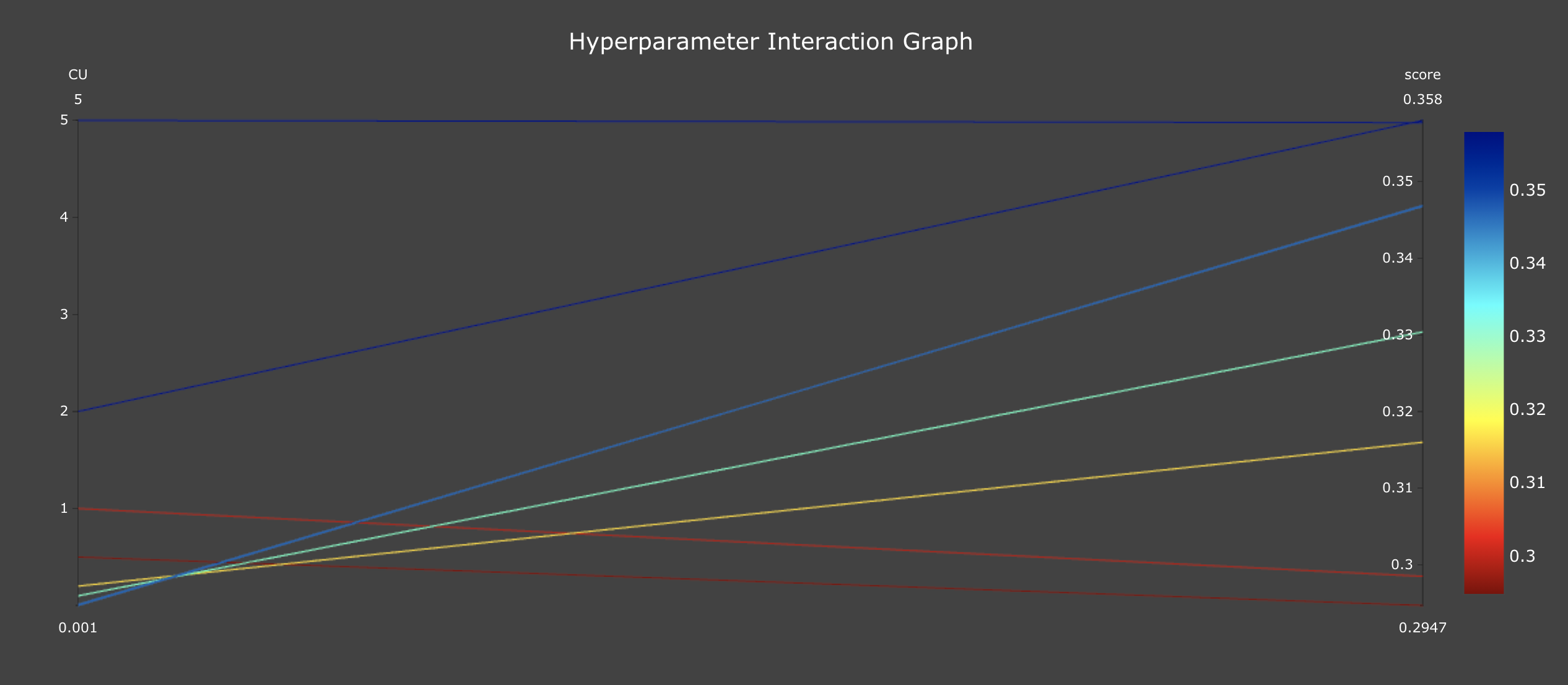}
\caption{SVHN performance based over different Cu values} \label{fig:svhn_cu} 
\end{figure*} 

\begin{figure*}[h]
\centering
\includegraphics[width=\textwidth]{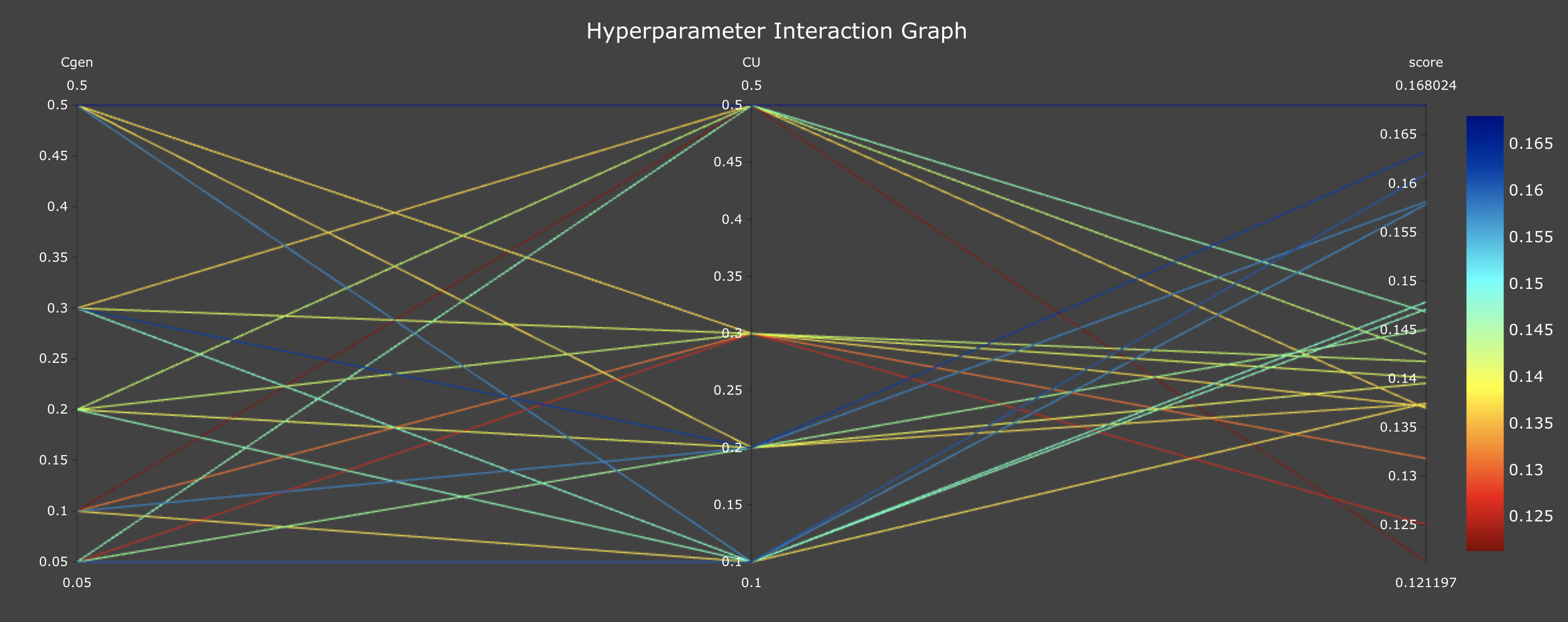}
\caption{SVHN performance over multiple Cu and Cgen values} \label{fig:svhn_both} 
\end{figure*} 

We analyze the performance of $C_U$ for SVHN discriminator accuracy, with $C_{gen}$ as 0 and $C_{gen}$ as another hyperparameter in Fig.~\ref{fig:svhn_cu} and Fig.~\ref{fig:svhn_both} respectively. For Fig.~\ref{fig:svhn_cu}, $C_U$ belongs to the set [5,2,1,0.5,0.2,0.1,0.01,0.001] and $C_{gen}$ is fixed at 0. For Fig.~\ref{fig:svhn_both}, we use a Random search on $C_U$ values [0.5,0.3,0.2,0.1] and with $C_{gen}$ values [0.5,0.3,0.2,0.1,0.05]. Based on our analysis of the hyperparameter interaction graphs in Figs. \ref{fig:svhn_cu} and \ref{fig:svhn_both}, we fix $C_U$ and $C_{gen}$ values of 0.5 and 0.1 respectively for our results in Table~\ref{tab:evolve}.

\subsection{CIFAR-10 data}

\begin{figure*}[h]
\centering
\includegraphics[width=\textwidth]{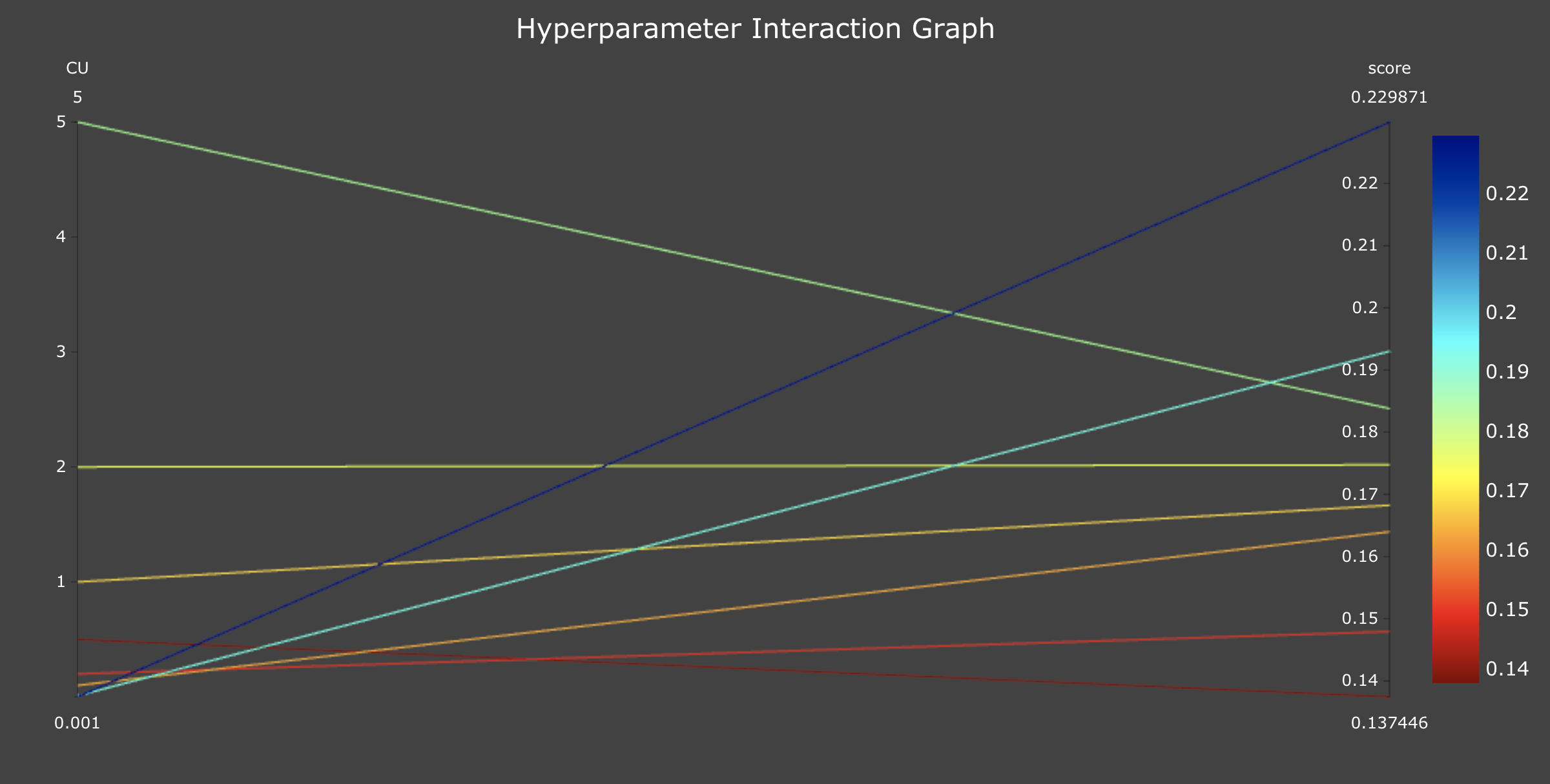}
\caption{Cifar performance based on Cu values} \label{fig:cif_cu} 
\end{figure*}

\begin{figure*}[h]
\centering
\includegraphics[width=\textwidth]{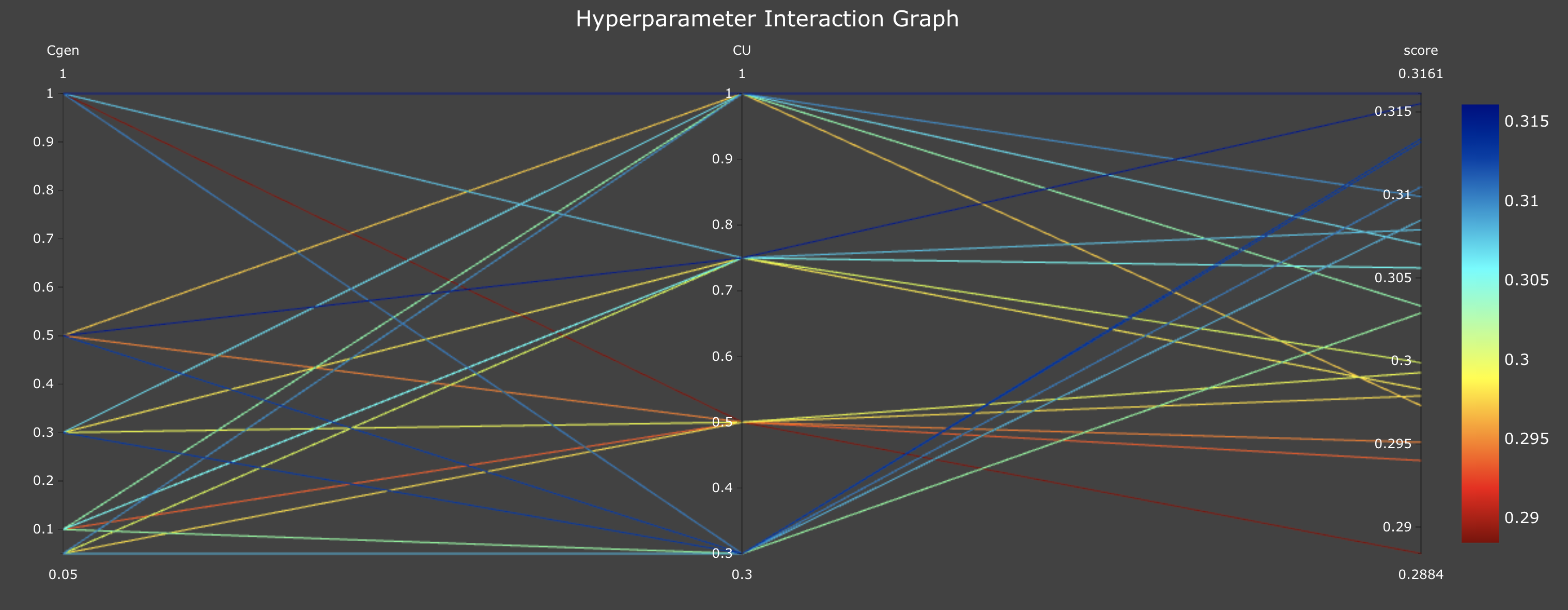}
\caption{Cifar performance based on Cu and Cgen values} \label{fig:cif_both} 
\end{figure*}

We analyze the performance of $C_U$ for Cifar-10 discriminator accuracy, with $C_{gen}$ as 0 and $C_{gen}$ as another hyperparameter in Fig.~\ref{fig:cif_cu} and Fig.~\ref{fig:cif_both} respectively. For Fig.~\ref{fig:cif_cu}, $C_U$ belongs to the set [5,2,1,0.5,0.2,0.1,0.01,0.001] and $C_{gen}$ is fixed at 0. For Fig.~\ref{fig:cif_both}, we use a Random search on $C_U$ values [1.0,0.75,0.5,0.3] and with $C_{gen}$ values [1.0,0.5,0.3,0.1,0.05]. Based on our analysis of the hyperparameter interaction graphs in Figs. \ref{fig:cif_cu} and \ref{fig:cif_both}, we fix $C_U$ and $C_{gen}$ values of 0.5 and 1.0 respectively for our results in Table~\ref{tab:evolve}. 

\subsection{Network Architecture}

Finally we provide the Discriminator and Generator architectures for the models we used for both the datasets SVHN and Cifar in Figs 12 - 15. These architectures are the same as used in \cite{dai2017good} and have been used for equivalent comparisons with baseline models. 

\begin{figure}[h]
\centering
\includegraphics[width=\textwidth]{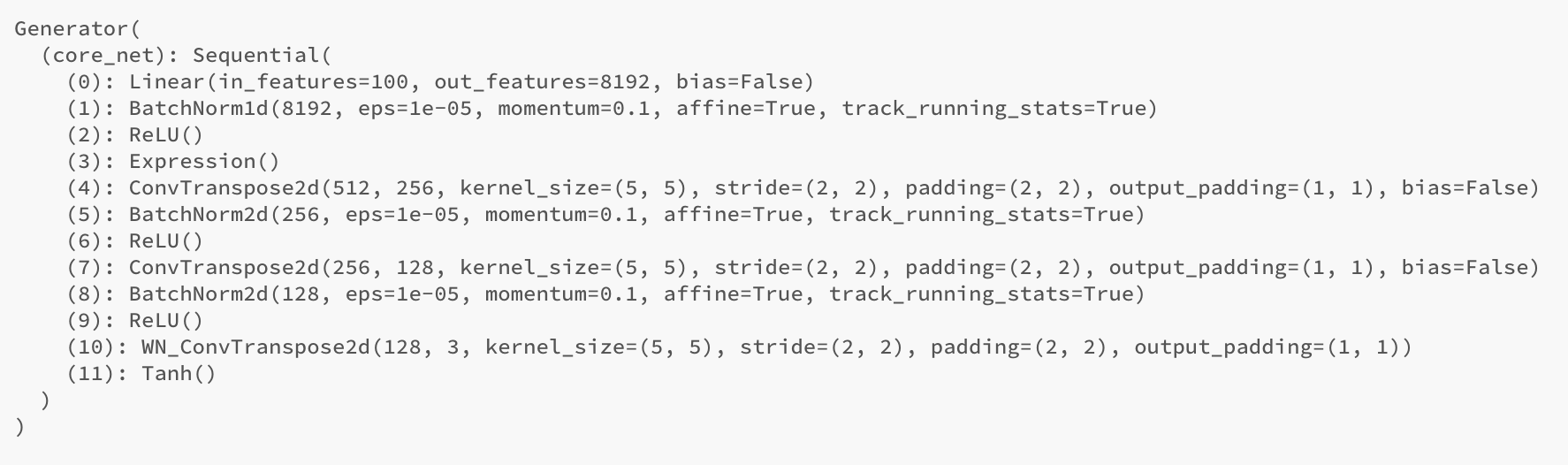}
\caption{SVHN Generator Architecture} \label{fig:svhn_gen_code} 
\end{figure} 

\begin{figure*}[h]
\centering
\includegraphics[width=\textwidth]{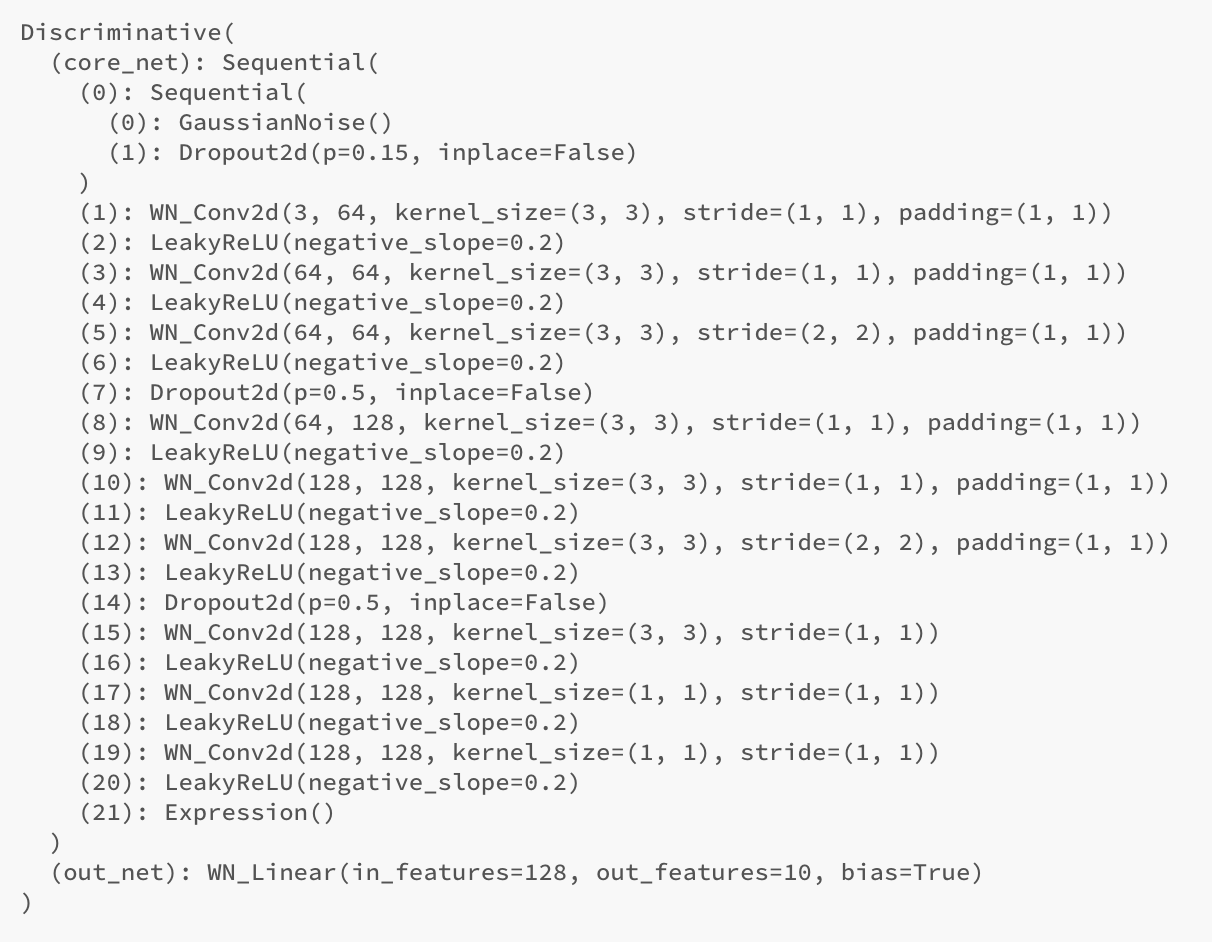}
\caption{SVHN Discriminator Architecture} \label{fig:svhn_dis_code} 
\end{figure*} 

\begin{figure*}[h]
\centering
\includegraphics[width=\textwidth]{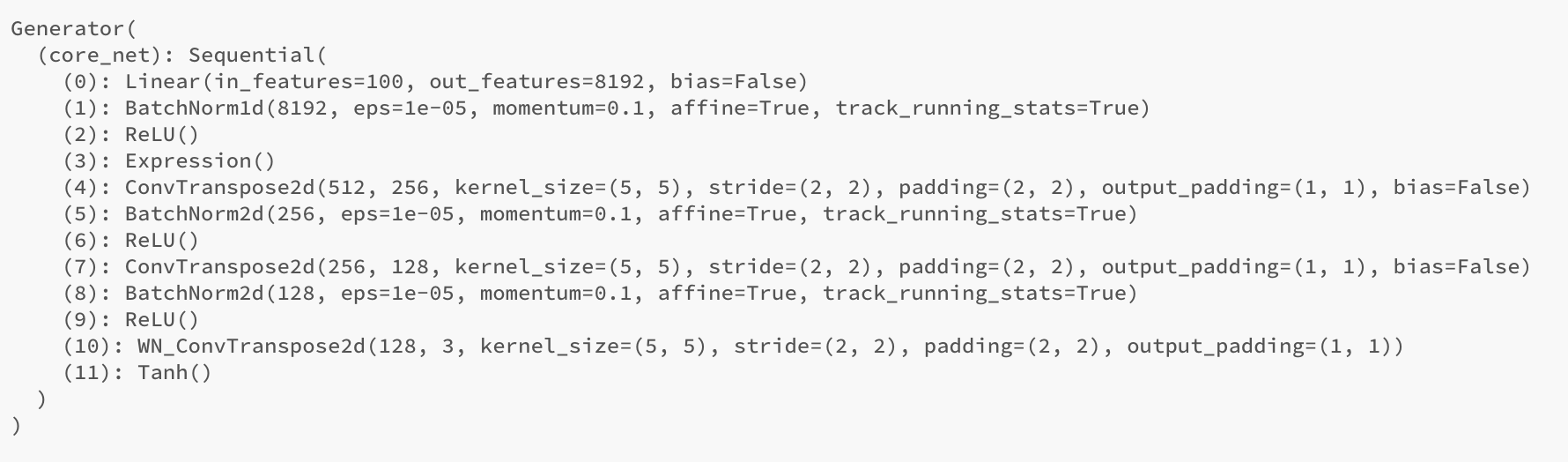}
\caption{Cifar Generator Architecture} \label{fig:cif_gen_code} 
\end{figure*} 

\begin{figure*}[h]
\centering
\includegraphics[width=\textwidth]{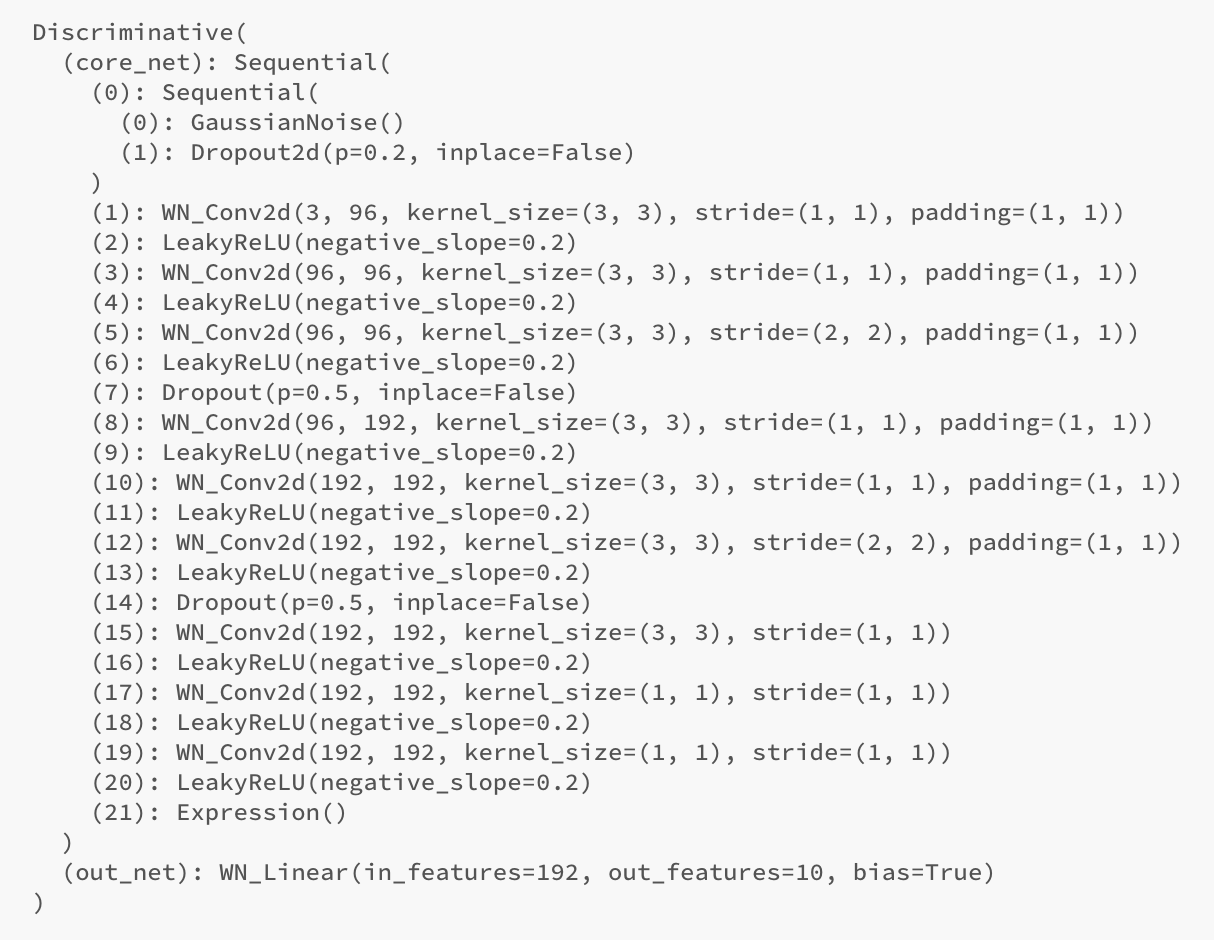}
\caption{Cifar Discriminator Architecture} \label{fig:cif_dis_code} 
\end{figure*} 

\clearpage
\subsection{U-GAN + VAT Experiment Setup} 
\label{app_UGAN_VAT}

In order to combine U-GAN with VAT, we first train the discriminator with the VAT loss instead of the universum one. When converged, we add the universum loss to the overall loss of the discriminator. For hyperparameters of the model, we use the optimal values we obtained for U-GAN and set the ones associated with the VAT model according to $\xi=10^{-5}, \epsilon=0.3, iteration=4, \alpha=30$~\cite{miyato2018virtual}.

\section{Future Research}
\label{sec:future_research}


There are two main directions for future research. \\
\noindent \textbf{Evolution Routine:} The current handling of the evolution process is hand-designed and may prove sub-optimal for different applications. A more systematic approach may be possible by connecting the existing theory in \cite{dhar2019} (Theorem 2) to transition from contradiction into compliance or by hyperparameter optimization of \texttt{epsSet} and \texttt{evolutionPeriod} using  tools like ~\cite{liu2019auptimizer,feurer2019auto}. Identifying the optimal transition (from contradiction to compliance) point and the evolution mechanism is paramount for the success of evolving GANs  and is an open research topic.

\noindent  \textbf{Extension to Advanced Learning Settings:} U-GAN and evolving GAN can be extended to other advanced learning techniques such as semi-supervised learning (similar to FM~\cite{salimans2016improved} and~C-GAN~\cite{dai2017good}). Such similar extensions to more advanced learning settings may yield additional performance improvements. Such extensions have not been explored in the current version and is a topic for future research.

\end{document}